\pdfoutput=1

\documentclass[11pt,a4paper]{article}
\usepackage[]{acl}

\usepackage{times}
\usepackage{latexsym}
\usepackage[T1]{fontenc}
\usepackage[utf8]{inputenc}
\usepackage{microtype}
\usepackage{xcolor}
\usepackage{adjustbox}
\usepackage{diagbox}
\usepackage{amsmath,amssymb}
\usepackage[inline]{enumitem}
\usepackage{array,multirow,graphicx}
\usepackage{tabularx}

\definecolor{mred}{RGB}{192, 0, 0}
\definecolor{mgreen}{RGB}{112, 173, 71}

\newcommand{\domaincount}{\textsc{D-con}}

\newcommand{\domainrel}{\textsc{d.rel}}
\newcommand{\labelpres}{\textsc{l.pres}}
\newcommand{\acceptability}{\textsc{accpt}}

\newcommand{\wer}{\textsc{wer}}

\newcommand{\model}{\texttt{DoCoGen}}
\newcommand{\modelf}{\texttt{F-DoCoGen}}

\newcommand{\dcrr}{\texttt{No-OV}}
\newcommand{\rcdr}{\texttt{RM-OV}}
\newcommand{\rcrr}{\texttt{RM-RR}}
\newcommand{\vae}{\texttt{VAE}}
\newcommand{\dann}{\texttt{DANN}}
\newcommand{\noda}{\texttt{NoDA}}
\newcommand{\eda}{\texttt{EDA}}
\newcommand{\oracle}{\texttt{Oracle-Gen}}
\newcommand{\perl}{\texttt{PERL}}
\newcommand{\perlaug}{\texttt{DoCoGen-PERL}}

\title{DoCoGen: Domain Counterfactual Generation for Low Resource Domain Adaptation}

\author{Nitay Calderon\Thanks{ Both authors equally contributed to this work.} \and Eyal Ben-David\footnotemark[\value{footnote}] \and Amir Feder \and Roi Reichart \\
  Technion - Israel Institute of Technology \\
{\tt \{nitay@campus.$|$eyalbd12@campus.$|$feder@campus.$|$roiri@\}technion.ac.il}
}

\date{}

\begin{document}
\maketitle

\begin{abstract}

Natural language processing (NLP) algorithms have become very successful, but they still struggle when applied to out-of-distribution examples. In this paper we propose a controllable generation approach in order to deal with this domain adaptation (DA) challenge. Given an input text example, our \model{} algorithm generates a domain-counterfactual textual example (\domaincount) --  that is similar to the original in all aspects, including the task label, but its domain is changed to a desired one. Importantly, \model{} is trained using only unlabeled examples from multiple domains -- no NLP task labels or parallel pairs of textual examples and their domain-counterfactuals are required. We show that \model{} can generate coherent counterfactuals consisting of multiple sentences.
We use the \domaincount s generated by \model{} to augment a sentiment classifier and a multi-label intent classifier in 20 and 78 DA setups, respectively, where source-domain labeled data is scarce. Our model outperforms strong baselines and improves the accuracy of a state-of-the-art unsupervised DA algorithm.\footnote{Our code and data are available at \url{https://github.com/nitaytech/DoCoGen}.}



\end{abstract}

\section{Introduction}
\label{sec:intro}

Natural Language Processing (NLP) algorithms are constantly improving and reaching significant milestones \citep{DBLP:conf/naacl/DevlinCLT19, DBLP:journals/jmlr/RaffelSRLNMZLL20, DBLP:conf/nips/BrownMRSKDNSSAA20}. However, such algorithms rely on the availability of sufficient labeled data and the assumption that the training and test sets are drawn from the same underlying distribution. Unfortunately, these assumptions do not hold in many cases due to the costly and labor-intensive data labeling process and since text may originate from many different domains. As generalization in low resource regimes and beyond the training distribution are still fundamental NLP challenges, NLP algorithms significantly degrade when applied to such scenarios.


Domain adaptation (DA) is an established field of research in NLP \citep{roark2003supervised, daume2006domain, DBLP:conf/acl/ReichartR07a} that attempts to explicitly address generalization beyond the training distribution (\S\ref{sec:related}). DA algorithms are trained on annotated data from source domains to be effectively applied in various target domains. Indeed, DA algorithms have been developed for multiple NLP tasks throughout the last two decades \citep{DBLP:conf/emnlp/BlitzerMP06, blitzer2007biographies, DBLP:conf/icml/GlorotBB11,
rush2012improved, DBLP:conf/conll/ZiserR17, DBLP:conf/emnlp/ZiserR18, ziser2018pivot, DBLP:conf/emnlp/HanE19}.



A natural alternative to costly human annotation would be to automatically generate labeled examples for model training. 
Doing so may expose the model to additional training examples and better represent the data distribution within and outside the annotated source domains. 
Unfortunately, generating labeled textual data is challenging \citep{feng2021survey}, especially when the available labeled data is scarce. Indeed, labeled data generation has hardly been applied to DA (\S\ref{sec:related}).



To allow DA through labeled data generation, we present \model , an  algorithm  that generates domain-counterfactual textual examples  (\domaincount s). In order to do that, \model{} intervenes on the domain-specific terms of its input example, replacing them with terms that are relevant for its target domain while keeping all other properties fixed, including the task label. Consider the task of sentiment classification (top example in Table~\ref{tab:examples}). When \model{} encounters an example from the \textit{Kitchen} domain (its source domain), it first recognizes the terms related to \textit{Kitchen} reviews, i.e., \textit{knife} and \textit{solid}. Then, it intervenes on these terms, replacing them with text that connects the example to the \textit{Electronics} domain (its target domain) while keeping the negative sentiment.

\model{} is a \textit{controllable generation} algorithm \citep{DBLP:conf/acl/LiGBSGD16, russo2020control} that is trained using a novel \textit{unsupervised} sentence reconstruction objective. Importantly, it does not require task-annotated data, or parallel pairs of sentences and their \domaincount s. 
A key component of \model{} is the \textit{domain orientation vector}, which guides the model to generate the new text in the desired domain. 
The parameters of the orientation vectors are learned during the unsupervised training process, allowing the generation model to share information among the various domains it is exposed to.



We focus on two low resource scenarios: Unsupervised domain adaptation (UDA) and any domain adaptation (ADA, \citet{ben2021pada}), with only a handful of labeled examples available from a single source domain. In both UDA and ADA the model is exposed to limited labeled source domain data and to unlabeled data from several domains. However, in UDA the \textit{unlabeled domains} contain the future target domain to which the model will be applied, while in ADA the model has no access to the target domain during training.
To cope with these extreme conditions, we use \model{} to enrich the source labeled data with \domaincount s from the unlabeled domains. 
By introducing labeled \domaincount s from various domains, we hope to provide the model with a training signal that is less affected by spurious correlations: Correlations between features and the task label which do not hold out-of-domain (OOD) \citep{DBLP:journals/corr/abs-2106-00545}.






After a brief evaluation of the intrinsic quality of the \domaincount s generated by \model, we evaluate our complete DA pipeline. We focus on two tasks: Binary sentiment classification of reviews and  multi-label intent prediction in information-seeking conversations.  
In both tasks, we follow the UDA and ADA scenarios, for a total of 12 and 8 sentiment setups, respectively, as well as 30 UDA and 48 ADA intent prediction setups.
Our results demonstrate the superiority of \model{}  over strong DA and textual-data augmentation algorithms. Finally, combining \model{} with PERL \citep{ben2020perl}, a SOTA UDA model, yields new SOTA DA accuracy and stability.



\begin{table*}
\centering
\footnotesize
\begin{adjustbox}{width=\textwidth}
\def\arraystretch{1.4}
\begin{tabularx}{\textwidth}{ X }

\hline
\hline

\underline{Original, \textbf{Kitchen}}: \sffamily{A good \textcolor{mred}{knife} but Quality Control was poor. The \textcolor{mred}{knife} is \textcolor{mred}{solid} and very comfortable in hand, however, when I got it new, the \textcolor{mred}{blade} is slightly \textcolor{mred}{bent}. I expect it to be in almost perfect \textcolor{mred}{condition}, but it's not.} \\

\underline{\model, \textbf{Kitchen $\rightarrow$ Electronics}}: \sffamily{A good \textcolor{mgreen}{product} but Quality Control was poor. The \textcolor{mgreen}{ipod} is \textcolor{mgreen}{very easy to use} and very comfortable in hand, however, when I got it new, the \textcolor{mgreen}{ipod} is slightly \textcolor{mgreen}{flimsy}. I expect it to be in almost perfect \textcolor{mgreen}{shape}, but it's not.} \\

\hline

\underline{Original, \textbf{DVD}}: \sffamily{The \textcolor{mred}{direction of} this \textcolor{mred}{film} is excellent. I love \textcolor{mred}{all the characters} and the way they interact. The \textcolor{mred}{storyline} is very important also. It's \textcolor{mred}{about religious beliefs} and neighbors that \textcolor{mred}{interact with} each other. It's a well-\textcolor{mred}{paced} and \textcolor{mred}{interesting story} that's not like anything else I've ever \textcolor{mred}{seen}.} \\

\underline{\model, \textbf{DVD $\rightarrow$ Airline}}: \sffamily{The \textcolor{mgreen}{service on} this \textcolor{mgreen}{flight} is excellent. I love \textcolor{mgreen}{the staff} and the way they interact. The \textcolor{mgreen}{safety} is very important also. It's \textcolor{mgreen}{nice to have staff} and neighbors that \textcolor{mgreen}{can help} each other. It's a well-\textcolor{mgreen}{groomed} and \textcolor{mgreen}{professional crew} that's not like anything else I've ever \textcolor{mgreen}{experienced}.} \\

\hline

\underline{Original, \textbf{Electronics}}: \sffamily{That \textcolor{mred}{relay board} is only good for \textcolor{mred}{switching AC loads} of \textcolor{mred}{100V} or more. If you have a lower \textcolor{mred}{voltage load}, it's not going to work. For low \textcolor{mred}{voltage loads} use \textcolor{mred}{transistors}, \textcolor{mred}{MOSFETs} or a \textcolor{mred}{ULN2803 driver board}.}
\\

\underline{\model, \textbf{Electronics $\rightarrow$ Statistics}}:  \sffamily{That \textcolor{mgreen}{model} is only good for \textcolor{mgreen}{data} of \textcolor{mgreen}{\$n\$} or more. If you have a lower \textcolor{mgreen}{\$n\$}, it's not going to work. For lower \textcolor{mgreen}{\$n\$ regression} use a \textcolor{mgreen}{linear}, \textcolor{mgreen}{logistic} or a \textcolor{mgreen}{t-test}.} \\

\hline
\hline
     
\end{tabularx}
\end{adjustbox}
\caption{Domain-counterfactual textual examples (\domaincount s) generated by \model{}. \textcolor{mred}{Red} terms are replaced with \textcolor{mgreen}{green} terms through the process of \domaincount{} generation. For additional examples see \S\ref{sec:examples}.}
\label{tab:examples}
\end{table*}

\section{Related Work}
\label{sec:related}
We first describe research in our DA setups: UDA and ADA. We then continue with the study of counterfactual-based data augmentation, and, finally, we describe research on counterfactual generation methods.

\paragraph{Domain Adaptation (DA)} 
The NLP literature contains several DA setups, the most realistic of which is \textit{unsupervised domain adaptation} (UDA), which assumes the availability of unlabeled data from a source and a target domain, as well as access to labeled data from the source domain \citep{DBLP:conf/emnlp/BlitzerMP06}. An even more challenging and potentially more realistic setup is the recently proposed \textit{any domain adaptation} setup (ADA, \citet{ben2021pada}), which assumes no knowledge of the target domains at training time. 
There are several approaches to DA, including  representation learning \citep{DBLP:conf/emnlp/BlitzerMP06, DBLP:conf/conll/ZiserR17} and data-centric approaches like instance re-weighting and self-training  \citep{DBLP:conf/nips/HuangSGBS06,  DBLP:journals/tacl/RotmanR19}.  

Since the rise of deep neural networks (DNNs), most focus in DA research has been directed to deep representation learning approaches (DReL). 
One line of DReL work employs an input reconstruction objective \citep{DBLP:conf/icml/GlorotBB11, DBLP:conf/icml/ChenXWS12, DBLP:conf/acl/YangE14, DBLP:journals/jmlr/GaninUAGLLML16}. Another line employs pivot features, which are prominent to the task of interest and common in the source and target domains \citep{blitzer2007biographies, DBLP:conf/www/PanNSYC10, ziser2018pivot, ben2020perl, lekhtman2021dilbert}.

We deviate from the DReL approach to DA and propose a data-centric methodology. Contrary to the above works, our approach can be applied to both UDA and ADA. Moreover, unlike previous ADA work, which builds upon multi-source DA, our approach can also perform single-source ADA.

\paragraph{Counterfactually Augmented Data (CAD)} 

Textual data augmentation (TDA) is a technique for increasing the training dataset without explicitly collecting new examples. This is achieved by adding slightly modified copies of already existing examples (local sampling) or newly created data (global sampling). TDA serves as a solution for insufficient data scenarios and as a technique for improving model robustness \citep{xie2020unsupervised, ng2020ssmba}. 
There are rule-based and model-based approaches to TDA. Rule-based methods commonly involve insertion, deletion, swap and replacement of specific words \citep{wei2019eda}, or template-based  paraphrasing \cite{rosenberg2021rad}. Model-based methods typically utilize a pretrained language model (PLM), e.g., for replacing random words \citep{kobayashi2018contextual, ng2020ssmba}, or generating entirely new examples from a prior data-distribution \citep{DBLP:conf/conll/BowmanVVDJB16, russo2020control, wang2021towards}. Other model-based methods apply backtranslation \citep{edunov2018understanding} or paraphrasing \citep{kumar2019submodular} for local sampling.


Another approach within local sampling TDA is to change (only) a specific concept that exists in the original example, creating a counterfactual example. Counterfactually-Augmented Data (CAD) is generated by minimally intervening on examples to change their ground-truth label, that is, perturbing only those terms necessary to change the label \citep{kaushik2019learning}. CAD is commonly used to improve generalizability \citep{kaushik2019learning, sen2021does}, however empirical results using CAD for OOD generalization have been mixed \citep{joshi2021investigation, khashabi2020more}. 


In this work, we explore a different type of counterfactuals, namely \domaincount s, which are the result of intervening only on the example's domain while holding everything else equal, particularly its task label. For sentiment analysis, we may be, for example, interested in revising a negative movie review, making it a negative airline review.
In addition, while CAD is mostly generated via a human-in-the-loop process \citep{kaushik2019learning, khashabi2020more, sen2021does}, our work focuses on automatic counterfactual generation.

\paragraph{Counterfactual Generation} 
\textit{controllable generation} refers to generation of text while controlling for specific attributes \citep{prabhumoye2020exploring}. 
The controlled attributes can range from style (e.g., politeness and sentiment) to content (e.g., keywords and entities) and even topic.
\citet{keskar2019ctrl} propose to control the generated text by training an LM on datasets annotated with the controlled attributes, and \citet{meister2020if} modify the model's decoding method. Recently, \citet{russo2020control} introduced a global sampling conditional variational autoencoder (VAE), augmenting text while controlling for attributes such as label and verb tense. However, controlling for the task label is challenging in scarce labeled data scenarios \citep{chen2021empirical}, since generative models require large amounts of labeled data .

Counterfactual generation lies at the intersection of controllable generation and causal inference \citep{feder2021causal}. Only few works deal with counterfactual generation, mostly by intervening on the task label. \citet{wu2021polyjuice} train a model on textual examples and their manually generated counterfactuals. Other works present methods for controlling for the text domain and semantics \citep{wang2020cat,feng2019keep}, yet they all experiment with short texts, while our model can generate longer texts, consisting of multiple sentences. A recent work by \citet{yu-etal-2021-cross} focuses on generation of new target-domain examples for aspect-based sentiment analysis (ABSA)  \citep{pontiki2016semeval}. However, this method is designed specifically for ABSA, utilizing predefined knowledge, and is only suitable for UDA setups where source domain labeled data is abundant. Our work presents a novel domain counterfactual generation algorithm, which can be trained in an unsupervised manner, and its generated outputs are demonstrated to be effective in multiple low-resource DA tasks.

\section{Domain-Counterfactual Examples}
\label{sec:dcon}

In this section, we formally define the concept of domain-counterfactual textual examples (\domaincount s) and discuss the motivation behind them.

\paragraph{Definition}

$x'$ is a \textit{domain-counterfactual example} (\domaincount) of $x$ if it is a coherent human-like text that is a result of intervening on the domain of $x$ and changing it to another domain, while holding everything else equal. 
Particularly, we would like the task label of $x'$ and $x$ to be identical. Formally, given an example $(x, y) \sim \mathcal{D}$ and a destination domain $\mathcal{D}'$, the goal of \domaincount{} generation is to generate $x' \sim P_{\mathcal{D}'}(X|Y=y)$ such that  $x' \simeq_\mathcal{D'} x$, where $\simeq_\mathcal{D'}$ is the domain counterfactual operator.

In this work, given a labeled source example $x$ we aim to generate coherent human-like \domaincount s from the unlabeled domains (see \S\ref{sec:intro}). We propose a \domaincount{} generation algorithm, \model{}, consisting of two components. The first involves masking domain specific terms of the given example, yielding $\texttt{M}(x)$. The second is a controllable generation model $\texttt{G}$ which takes as input $\texttt{M}(x)$ and a \textit{domain orientation vector} $v'$. This vector specifies the destination domain $\mathcal{D}'$, controlling the semantics of the generated \domaincount . Formally:

\vspace{-1.25em}

\begin{equation*}
    \model(x, \mathcal{D}') = \texttt{G}(\texttt{M}(x), v') \simeq_\mathcal{D'} x
\end{equation*}

\paragraph{Motivation}
The NLP community has recently become increasingly concerned with \emph{spurious correlations}~\cite{geirhos2020shortcut,wang-culotta-2020-identifying,gardner2021competency}. In the case of DA, spurious correlations may be defined as correlations between $X$ and $Y$ which are relevant only to a specific domain or in a certain sample of labeled examples. 
Such correlations may make a predictor $f: \mathcal{X} \to \mathcal{Y}$ brittle to domain shifts.


Using counterfactuals w.r.t. a specific variable allows us to both estimate its effect on our predictor \cite{feder2021causalm, rosenberg2021rad} or alleviate its impact on it \cite{kaushik2020explaining}. We focus on the latter, automatically generating \domaincount s by intervening on the domain variable $\mathcal{D}$. Adding these \domaincount s to the training set of a predictor should reduce its reliance on domain-specific information and spurious correlations.

From a DA perspective, enriching the training data with \domaincount s is motivated by pivot features (\S\ref{sec:related}), which are frequent in multiple domains and are prominent for the task. \domaincount s preserve language patterns, such as pivots, which are frequent in multiple domains. 
Consider the middle example in Table~\ref{tab:examples}, pivot words (such as \textit{excellent} and \textit{important}) are preserved in the \domaincount , while non-pivots (\textit{intereseting} and \textit{well-paced}) are replaced due to the domain intervention. 
 Accordingly, a model trained on an example and its \domaincount{} is directed to focus on pivots rather than on non-pivots, consequently generalizing better OOD.

\begin{figure*}[t]
    \centering
    \includegraphics[width=0.95\textwidth]{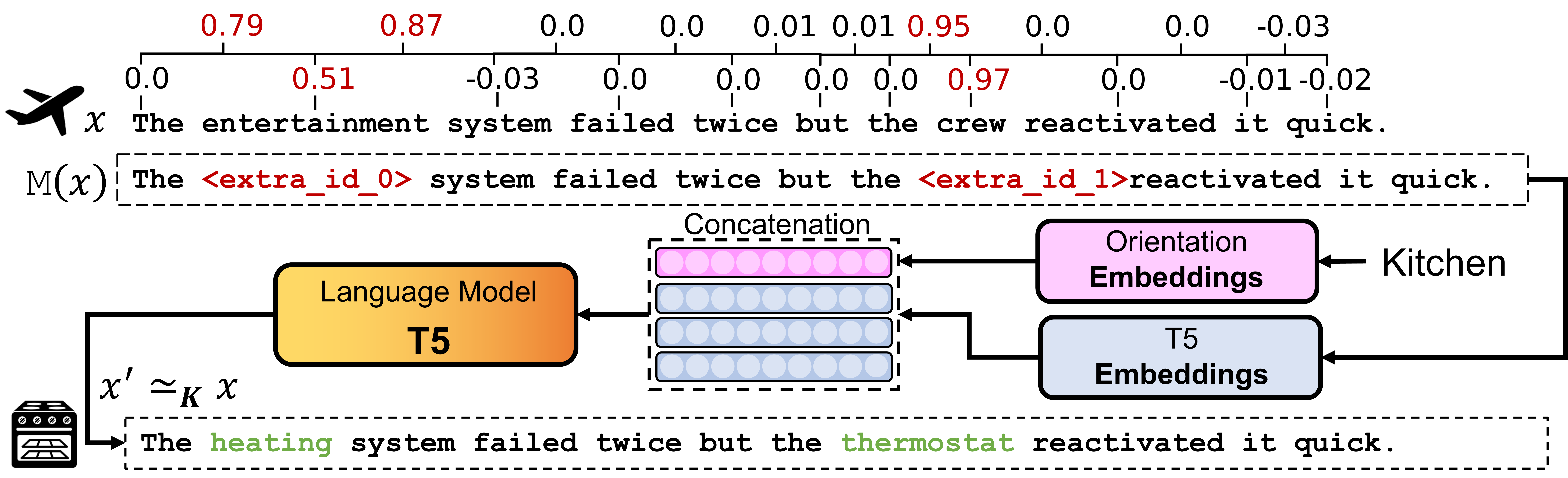}
    \caption{The \model{} model. Given a review $x$ from the \textit{airline} domain, we aim to generate a \domaincount{} from the \textit{kitchen} domain. We first corrupt the domain of the example by masking domain specific terms. The numbers above the input words are the masking scores of uni-grams and bi-grams. Terms with scores above a threshold ($\tau=0.08$) are masked. 
    In the reconstruction step we use a T5-based generation model to generate the \domaincount{} $x' \simeq_{\textbf{K}} x$. The input of the model is a concatenation of the \textit{orientation vector} that represents the target domain with the model's embedding vectors which correspond to the tokens of the masked example $\texttt{M}(x)$.}
    \label{fig: model}
\end{figure*}

\section{DoCoGen: Domain Counterfactual Generation}
\label{sec:model}

We propose a corrupt-and-reconstruct approach  for generating \domaincount s from given source domain examples (Figure~\ref{fig: model}).
We next extend on these two steps, and describe our filtering mechanism used to disqualify low quality \domaincount s.



\subsection{Domain Corruption}
\label{sub:domain-coruption}

The first step of generating a \domaincount{} is to mask domain specific terms. In order to mask an example $x \sim \mathcal{D}$ with a destination domain $\mathcal{D}'$, we first mask all uni-grams $w$ with $\texttt{m}(w,\mathcal{D}, \mathcal{D}')>\tau$, where $\tau$ is a hyperparameter and $\texttt{m}$ is a masking score that is defined later in this section. Then, we mask all the remaining bi-grams (that do not contain a masked uni-gram) according to the same masking threshold $\tau$. 
This process is repeated up to tri-gram expressions. 
The final output of the corruption step is a masked example $\texttt{M}(x)$.

In Figure~\ref{fig: model}, the masking scores of  uni-grams and bi-grams appear above the input words. An n-gram is masked if and only if its score is above a $\tau = 0.08$ threshold and the scores of its grams are lower. For example, \textit{system} is not masked although the bi-gram \textit{entertainment system} has a score above the $\tau$ threshold, since \textit{entertainment} is masked and the score of \textit{system} is lower than $\tau$. 

\paragraph{Masking Score}
Let $w$ be an n-gram and $\mathcal{D}$ be a domain with $n_\mathcal{D}$ unlabeled examples. We denote the number of examples from $\mathcal{D}$ that contain $w$ by $\#_{w|\mathcal{D}}$. By assuming that domains have equal prior probabilities and by using the Bayes' rule, the probability of $\mathcal{D}$ given $w$ can be estimated by $P(D=\mathcal{D}|W=w) \propto \frac{\#_{w|\mathcal{D}} + \alpha}{n_\mathcal{D}}$, where $\alpha$ is a smoothing hyperparameter.
We define the affinity of $w$ to  $\mathcal{D}$ to be:
$$\rho(w, \mathcal{D})= P(\mathcal{D}| w) \cdot \bigg(1-\frac{H(D|w)}{\log{N}}\bigg)$$ 
where $N$ is the number of unlabeled domains and $H(D|w)$ is the entropy of $D|w$, which is upper bounded by $\log{N}$. Notice that higher $H(D|w)$ values indicate that $w$ is not related to any specific domain.
Finally, we set the masking score of an n-gram $w$ with an origin domain $\mathcal{D}$ and a destination domain $\mathcal{D}'$ as follows:
\begin{equation*}
    \texttt{m}(w, \mathcal{D}, \mathcal{D}') = \rho(w, \mathcal{D}) - \rho(w, \mathcal{D}')
\end{equation*}

Note that $\texttt{m}(w, \mathcal{D}, \mathcal{D}') \in [-1, 1]$. It can be negative due to the right hand side's subtrahend, which aims to prevent masking n-grams that are related to the destination domain and should appear in the counterfactual, like \textit{system} in Figure~\ref{fig: model}.  

\subsection{Domain-Oriented Reconstruction}
\label{sub:oriented-reconstruction}

The second step of \model{} is a reconstruction step that involves a generative model, based on an encoder-decoder T5 architecture \cite{DBLP:journals/jmlr/RaffelSRLNMZLL20}. Given a masked example $\texttt{M}(x)$ and a destination domain $\mathcal{D}'$, we concatenate a domain orientation vector $v'$ that represents $\mathcal{D}'$ with the masked input's embedding vectors. Then, the concatenated matrix is passed as an input to the encoder-decoder model for counterfactual generation, yielding $x'$. We next describe the mechanism behind domain orientation vectors.

\paragraph{Domain Orientation Vectors}
In addition to the T5 embedding matrix (T5 Embeddings in Figure~\ref{fig: model}), we equip our model with another learnable embedding matrix, containing $K \cdot N$ orientation vectors, such that each domain is represented by $K$ different vectors (Orientation Embeddings in Figure~\ref{fig: model}). 
We initialize the orientation vectors with the T5 embedding vectors of the domain names and the top $K-1$ \textit{representing words} of each domain. The top representing words of domain $\mathcal{D}$ are those which reach the highest score of: $\log{(\#_{w|\mathcal{D}}+1)}\rho(w, \mathcal{D})$. We use $K$ orientation vectors to allow us generate a heterogeneous set of \domaincount s for a given destination domain (see examples in \S\ref{sec:examples}). We note that although the orientation vectors are initialized with vectors from the T5 embedding matrix, they have a different role and thus are likely to converge to different values during the training process.

\paragraph{Training} In the spirit of low resource learning, we would like to train \model{} in an unsupervised manner, i.e., without access to manually generated \domaincount s. Therefore, we use the unlabeled data of our unlabeled domains. For each example $x$, we provide the model with $\texttt{M}(x)$, the corrupted version of $x$, and $v$, the orientation vector of $\mathcal{D}$, and with $x$ as the gold output. The model hence learns to reconstruct $x$ given $\texttt{M}(x)$ and $v$.

Notice that the origin and the destination domains are the same, i.e, $\mathcal{D} = \mathcal{D}'$, and the masking score is $\texttt{m}(w, \mathcal{D}, \mathcal{D})=0$. Hence, for masking purposes, we randomly choose $\Tilde{\mathcal{D}} \ne \mathcal{D}$ and plug it as the destination domain in the masking score. We then choose an orientation $v$ for $\mathcal{D}$, by randomly sampling either the domain name or one of its representing words as long as it appears in $x$.

Finally, since the orientation vector parameters are trained as part of the reconstruction objective, 
we establish the connection between the orientation vector and the semantics of the completed example. Hence, we expect that at inference time examples will be properly transformed into their \domaincount s. 


\paragraph{Inference} Given $(x, \mathcal{D}, \mathcal{D}')$, we first mask the example to get $\texttt{M}(x)$ and select one orientation vector $v'$ that represents $\mathcal{D'}$.\footnote{\S\ref{sub:masking} presents the $\%$ of masked tokens in our experiments.} Together, the tuple $(\texttt{M}(x), v')$ forms the input, and accordingly the model generates a \domaincount{} $x' \simeq_\mathcal{D'} x$. To increase the likelihood that $x'$ originates from $\mathcal{D}'$, we restrict the model to generate only tokens of the original example or tokens that are related to $\mathcal{D}'$ and meet the  condition:  $\max_{i\in{1,..., N}} \texttt{m}(w, \mathcal{D}', \mathcal{D}_i)>\tau$.

\subsection{Filtering Mechanism}
\label{sub:filtering-mech}

In order to properly apply \model{} within a DA pipeline, we introduce a filtering mechanism that disqualifies low quality \domaincount s generated by \model. Particularly, we train a classifier to predict the domain of the original, human-written unlabeled examples, and use it to remove \domaincount s if their predicted domain is not the given destination domain. 
In addition, we disqualify \domaincount s with less than four words or when the word overlap with the original example is lower than $25\%$. 
We name \model{} when equipped with this filtering mechanism  \modelf . 

\section{Intrinsic Evaluation}
\label{sec:intrinsic}

We next assess \model{} in terms of its generated \domaincount s, ensuring they:
\begin{enumerate*}[label=(\roman*), itemjoin={{, }}, itemjoin*={{, and }}]
  \item belong to the correct domain and label (1, 2)
  \item are fluent (3, 4).
\end{enumerate*}
To this end, we collected 20 original reviews, equally distributed among four domains (the A, D, E, and K domains, see \S\ref{sec:exp-setup}). We then applied \model{} to generate 60 \domaincount s, 3 for each of the original reviews (see \S\ref{sec:exp-setup} for the \model{} training setup). Finally, we trained the \vae{} model of \citet{russo2020control} on labeled data (all the labeled data of the A, D, E, and K domains) and applied it to generate five reviews from each of the above four domains, with the same number of positive and negative reviews as in the set of original reviews.

We then conducted a crowd-sourcing experiment where five nearly native English speakers rated each example, considering the following evaluation measures:
(1) Domain relevance (\domainrel) - whether the topic of the generated text is related to its destination domain; (2) Label preservation (\labelpres) - what is the label of the generated example (and we report whether the answer was identical to the desired label); (3) Linguistic Acceptability (\acceptability) - how logical and grammatical the example is (on a 1-5 scale); and 
(4) Word error rate (\wer) - what is the minimum number of word substitutions, deletions, and insertions that have to be performed to make the example logical and grammatical.\footnote{We actually asked the annotators to edit the example and then measured the number of edit operations.}

Table~\ref{tab:hum-intrinsic} reports our results. 
\model{} achieves high \acceptability{} scores and low \wer{} scores, significantly outperforming its \vae{} alternative, which is known to struggle with longer texts \citep{shen2020towards, iqbal2020survey}. Interestingly, \model{} achieves compatible results to the original reviews, indicating the high quality of its generated texts.
Finally, in more than $90\%$ of the cases \model{} manages to change the example domain to the desired domain, and in $80\%$ it preserves the original example label. In comparison, only $88\%$ of the original examples were annotated as their gold label.

\begin{table}
\small
\centering
\begin{adjustbox}{width=0.48\textwidth}

\begin{tabular}{ | l || c | c | c | c |}

\hline

 & \textbf{$\uparrow$\domainrel } & \textbf{$\uparrow$\labelpres} & \textbf{$\uparrow$\acceptability} & \textbf{$\downarrow$\wer} \\
\hline
 \vae & $ 90.0 $ &  $ 46.0 $ & $ 2.11 $ & $ 0.54 $ \\

 \model & $ 93.0 $ &  $ 80.0 $ & $ 4.01 $ & $ 0.17 $ \\

 \texttt{Original Reviews} & $ 99.0 $ &  $ 88.0 $ & $ 4.73 $ & $ 0.10 $ \\
\hline

\end{tabular}
\end{adjustbox}
\caption{Human intrinsic evaluation. Up arrows ($\uparrow$) represent metrics where higher scores are better, and down arrows ($\downarrow$) represent the opposite. }
\label{tab:hum-intrinsic}
\end{table}



  




\section{Experimental Setup}
\label{sec:exp-setup}

\subsection{Tasks and Domains\footnote{URLs of the datasets and the code, implementation and hyperparameter details are described in \S\ref{sub:implementation}.}}


In this subsection we describe our tasks and datasets, as well as the two DA setups which are the focus of this work. A full description of the number of samples in each dataset is found in Table~\ref{tab:datasets}.

\paragraph{Sentiment Classification}

We follow a large body of prior DA work, focusing on the task of binary sentiment classification. Specifically, our experiments include six different domains: the four legacy product review domains \citep{blitzer2007biographies} - Books (B), DVDs (D), Electronic items (E) and Kitchen appliances (K); the challenging airline review dataset (A) \cite{Nguyen2015airline, ziser2018pivot}; and the restaurant (R) domain obtained from the Yelp dataset challenge \citep{DBLP:conf/nips/ZhangZL15}.
The focus of this work is on low resource DA, and thus we randomly sample 100 labeled examples to form the training set for the following domains: A, D, E, and K.

As described in \S\ref{sec:related}, we explore two DA setups, UDA and ADA. For UDA, where the model has access to unlabeled target domain data, we experiment with 12 cross-domain setups, including the following domains: A, D, E, and K. For ADA, where unlabeled data from the target domain is not within reach, we experiment with a total of 8 setups, including B and R as target domains, and A, D, E, and K as source domains. Our reported accuracy scores are averaged across 25 different seeds and randomly sampled training and development sets.

\paragraph{Multi-label intent prediction}

Our second task is multi-label intent prediction of utterances from information-seeking conversations. We use the multi-domain MANtIS dataset \citep{DBLP:journals/corr/abs-1912-04639}, consisting of diverse conversations from the question-answering Stack Exchange portal. The authors provide manually annotated user intent utterances, with eight possible intent labels, such as \textit{information request}, \textit{potential answer} and \textit{greetings}. Since we focus on low resource scenarios, we use only the five most common labels, as the frequency of the other three labels is less than 5\%, and in some domains they are completely missing.

The MANtIS dataset consists of 14 domains: Apple (AP), DBA (DB), Electronics (EL), Physics (PH), Statistics (ST), askubuntu (UB);  DIY (DI), English (EN), Gaming (GA), GIS (GI), Sci-Fi (SC), Security (SE), Travel (TR) and Worldbuilding (WO). We use the first 6 domains as unlabeled domains, randomly sampling train, development and test sets for each. The remaining 8 domains are used as target domains in the ADA setup, resulting in 30 UDA ($6 \times 5$) and 48 ADA ($6 \times 8$) setups.

Following \citet{DBLP:journals/corr/abs-1912-04639}, we use the (Macro) F1-score to measure classifier performances, and, like in the sentiment classification task, our reported results are averaged across 25 different seeds and randomly sampled training sets.

\paragraph{DA by Augmentation} 

The DA pipeline includes a T5-based sentiment classifier trained on labeled data from a single source domain and an augmentation model (e.g., \model) trained on unlabeled data from four unlabeled domains. We first train \model{} on the unlabeled data, and then use it for generating \domaincount s that enrich the classifier's training data. For each labeled training example, \model{} generates $K=4$ \domaincount s w.r.t. each unlabeled domain, resulting in a total of $16$ \domaincount s per example. After training the sentiment classifier on the enriched data, we evaluate it on test examples originating from one of the unlabeled domains (UDA) or one of the unseen domains (ADA). We denote each DA model by the algorithm that was used for enriching its training data.

\subsection{Models and Baselines}
\label{sub:models-and-baselines}


Our main models are \model{} and \modelf , which is equipped with the filtering mechanism.
We compare them to three types of models: (a) baseline models, including both baselines for the entire DA pipeline (1,2,5) and 
alternative augmentation methods (3,4);
(b) ablation models (6,7) that use variants of our \domaincount{} generation algorithm where one component is modified, highlighting the importance of our design choices; and (c) an upper-bound generation model that has access to labeled data from the target domains. Unless otherwise stated, all sentiment classifiers use the same architecture, based on a pre-trained T5 model. We next describe the models in each of these groups.


\paragraph{Baseline DA Models}

We experiment with five baselines: (1) \textit{No-Domain-Adaptation} (\noda),
A model that is only trained on the available training data from the source domain in each DA setup; (2) \textit{Domain-Adversarial-Neural-Network} (\dann), 
A model that integrates the sentiment analysis predictive task with an adversarial domain classifier to learn domain invariant representations \citep{DBLP:journals/jmlr/GaninUAGLLML16}. This model does not apply augmentation, but instead the unlabeled data is used for training its adversarial component; (3) \textit{Easy-Data-Augmentation (\eda)}, an augmentation method that randomly inserts, swaps, and deletes words or replaces synonyms \citep{wei2019eda}; (4) \textit{Random-masking Random-Reconstructing} (\rcrr), another basic augmentation method that randomly masks tokens from the input example and then fills the masks with tokens that are chosen by a masked language modeling head, as suggested by \citep{ng2020ssmba}; and (5) \textit{PERL}, a SOTA model for the UDA setup \citep{ben2020perl}.

\begin{table*}
\centering
\begin{adjustbox}{width=\textwidth}

\begin{tabular}{ | l || c | c | c | c | c | c |c | c | c | c | c | c || c | }

\hline
  & \textbf{A $\rightarrow$ D} & \textbf{A $\rightarrow$ E} & \textbf{A $\rightarrow$ K} & \textbf{D $\rightarrow$ A} & \textbf{D $\rightarrow$ E} & \textbf{D $\rightarrow$ K} & \textbf{E $\rightarrow$ A} & \textbf{E $\rightarrow$ D} & \textbf{E $\rightarrow$ K} & \textbf{K $\rightarrow$ A} & \textbf{K $\rightarrow$ D} & \textbf{K $\rightarrow$ E} & \textbf{AVG} \\
  \hline
  
	\noda & $69.4$ & $78.6$ & $78.2$ & $72.3$ & $80.2$ & $82.4$ & $81.0$ & $79.8$ & $87.6$ & $72.5$ & $78.6$ & $85.4$ & $78.8$ \\

	\dann & $70.3$ & $78.7$ & $78.9$ & $75.5$ & $81.2$ & $82.3$ & $82.3$ & $78.3$ & $86.7$ & $81.0$ & $78.3$ & $85.0$ & $79.9$ \\

	\eda & $69.3$ & $79.1$ & $79.4$ & $71.1$ & $79.9$ & $83.0$ & $79.9$ & $80.8$ & $88.0$ & $75.7$ & $80.9$ & $86.4$ & $79.5$ \\

	\rcrr & $69.5$ & $80.1$ & $80.0$ & $72.3$ & $81.0$ & $83.8$ & $79.6$ & $79.5$ & $88.4$ & $70.6$ & $79.1$ & $84.5$ & $79.0$ \\ 

\hline

	\dcrr & $67.2$ & $76.5$ & $76.1$ & $71.5$ & $79.7$ & $82.9$ & $80.9$ & $80.5$ & $88.9$ & $74.8$ & $79.6$ & $85.3$ & $78.7$ \\

	\rcdr & $69.3$ & $\textbf{80.2}$ & $\textbf{80.4}$ & $72.7$ & $81.8$ & $84.5$ & $79.6$ & $81.7$ & $89.0$ & $70.3$ & $79.4$ & $85.4$ & $79.5$ \\ 

\hline

	\model & $70.6$ & $79.7$ & $79.8$ & $75.8$ & $82.8$ & $84.4$ & $\textbf{83.0}$ & $82.0$ & $\textbf{89.3}$ & $81.2$ & $82.2$ & $87.3$ & $81.5$ \\

	\modelf & $\textbf{71.1}$ & $79.6$ & $79.6$ & $\textbf{76.7}$ & $\textbf{83.2}$ & $\textbf{84.8}$ & $82.6$ & $\textbf{82.1}$ & $89.2$ & $\textbf{81.4}$ & $\textbf{83.3}$ & $\textbf{88.0}$ & $\textbf{81.8}$ \\ 

\hline
\hline
  \perl & $72.9$ & $81.1$ & $\underline{83.6}$ & $81.5$ & $83.0$ & $\underline{86.9}$ & $81.1$ & $\underline{81.7}$ & $\underline{88.5}$ & $77.9$ & $78.2$ & $86.1$ & $81.9$ \\
  
  \perlaug & $\underline{75.7}$ & $\underline{82.7}$ & $83.1$ & $\underline{82.4}$ & $\underline{85.0}$ & $84.9$ & $\underline{81.3}$ & $80.8$ & $88.3$ & $\underline{79.5}$ & $\underline{80.9}$ & $\underline{86.2}$ & $\underline{82.6}$ \\

\hline
\hline

	\oracle & $83.8$ & $88.4$ & $88.9$ & $83.6$ & $89.3$ & $90.0$ & $84.9$ & $84.6$ & $90.7$ & $84.1$ & $82.2$ & $89.0$ & $86.6$ \\ 

\hline

\end{tabular}
\end{adjustbox}
\caption{Sentiment classification: accuracy scores for each source and target domain pair in the UDA setup. \textbf{Bold} numbers mark the best performing T5-based model, and \underline{underline} numbers mark the best performing \perl{} model.}
\label{tab:uda_reviews}
\end{table*}

\begin{table}
\centering
\begin{adjustbox}{width=0.48\textwidth}

\begin{tabular}{ | l || c | c | c | c | c | c | c | c || c | }
\hline
  Source & \multicolumn{2}{c|}{A} & \multicolumn{2}{c|}{D} & \multicolumn{2}{c|}{E} & \multicolumn{2}{c||}{K} &  \\

\hline
  Target & \textbf{B} & \textbf{R} & \textbf{B} & \textbf{R} & \textbf{B} & \textbf{R} & \textbf{B} & \textbf{R} & \textbf{AVG} \\
  \hline
  \hline
  
	\noda & $69.1$ & $76.5$ & $82.3$ & $82.8$ & $81.5$ & $84.5$ & $82.4$ & $85.2$ & $80.5$ \\

	\dann & $70.5$ & $77.2$ & $82.7$ & $81.5$ & $80.9$ & $83.4$ & $81.8$ & $83.4$ & $80.2$ \\

	\eda & $69.3$ & $78.0$ & $83.7$ & $82.6$ & $83.2$ & $85.4$ & $82.8$ & $86.3$ & $81.4$ \\

	\rcrr & $69.4$ & $78.4$ & $83.8$ & $83.5$ & $81.9$ & $85.6$ & $83.7$ & $85.4$ & $81.5$ \\ 

\hline

	\dcrr & $67.1$ & $76.1$ & $83.8$ & $82.5$ & $82.9$ & $\textbf{86.2}$ & $83.0$ & $85.6$ & $80.9$ \\

	\rcdr & $69.6$ & $78.7$ & $84.3$ & $\textbf{83.6}$ & $83.6$ & $\textbf{86.2}$ & $83.9$ & $85.5$ & $81.9$ \\ 

\hline

	\model & $70.9$ & $78.1$ & $84.4$ & $82.9$ & $83.9$ & $86.0$ & $84.5$ & $85.7$ & $82.1$ \\

	\modelf & $\textbf{71.4}$ & $\textbf{79.3}$ & $\textbf{84.9}$ & $\textbf{83.6}$ & $\textbf{84.2}$ & $86.1$ & $\textbf{85.6}$ & $\textbf{87.2}$ & $\textbf{82.8}$ \\ 

\hline
\hline

	\oracle & $84.4$ & $85.2$ & $86.7$ & $86.1$ & $86.0$ & $86.5$ & $85.3$ & $86.5$ & $85.8$ \\
	
\hline

\end{tabular}
\end{adjustbox}
\caption{Sentiment classification: accuracy scores for each source and target domain pair in the ADA setup. }
\label{tab:ada_reviews}
\end{table}

\paragraph{Ablation Models}

We consider two variants of \model: 
(6) \textit{No-Orientation-Vectors} (\dcrr),
a generation model that masks tokens by employing a similar masking mechanism as \model, and then employing a masked language modeling head to fill the masked tokens (without domain orientation vectors); and (7) \textit{Random-Masking with Orientation-Vectors} (\rcdr), a generation model that randomly masks tokens from the input example and then employs the \model 's reconstruction mechanism to fill the masks.

\paragraph{Upper-Bound}
We implement an upper-bound model for \domaincount{} augmentation, \textit{Oracle-Matching} (\oracle). Unlike all other models in this work, \oracle{} has access to target domain labeled data. Thus, given an example from a source domain, \oracle{} looks for the most similar example with the same label in the target domain, and adds it to its training data (see \S\ref{sub:urls}).

\begin{figure}
    \centering
    \includegraphics[width=0.48\textwidth]{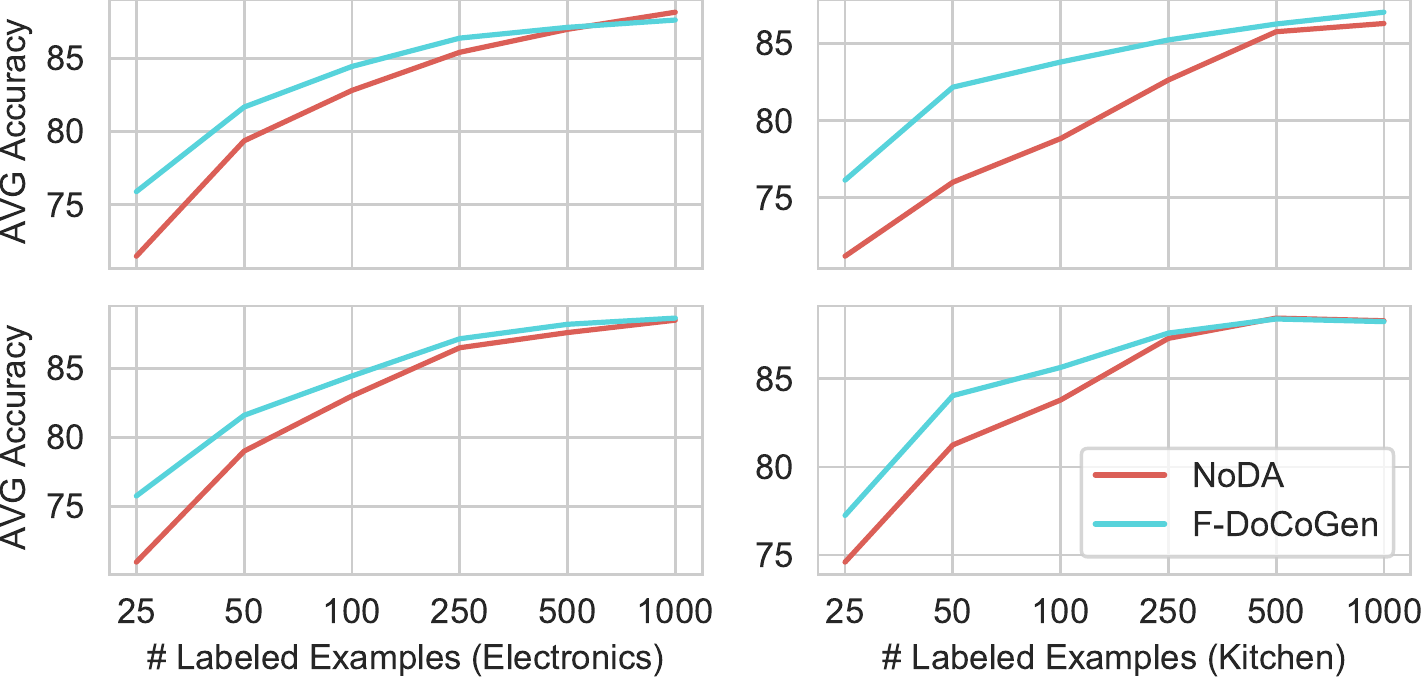}
    \caption{Average accuracy in UDA (top) and ADA (bottom) setups with different number of labeled examples from two source domains: E and K.}
    \label{fig:labeled_size}
\end{figure}

\begin{table*}
\centering
\begin{adjustbox}{width=\textwidth}

\begin{tabular}{ | l || c | c | c | c | c | c | c | c | c | c | c | c || c | c | }
\hline
  Source & \multicolumn{2}{c|}{AP} & \multicolumn{2}{c|}{DB} & \multicolumn{2}{c|}{EL} & \multicolumn{2}{c|}{PH} & \multicolumn{2}{c|}{ST} & \multicolumn{2}{c||}{UB} & \multicolumn{2}{c|}{AVG} \\

\hline
  Setup & \textbf{UDA} & \textbf{ADA} & \textbf{UDA} & \textbf{ADA} & \textbf{UDA} & \textbf{ADA} & \textbf{UDA} & \textbf{ADA} & \textbf{UDA} & \textbf{ADA} & \textbf{UDA} & \textbf{ADA} & \textbf{UDA} & \textbf{ADA} \\
  \hline
  \hline
  
	\noda & $75.5$ & $74.3$ & $72.2$ & $71.0$ & $71.2$ & $70.8$ & $67.1$ & $67.0$ & $71.8$ & $70.0$ & $72.0$ & $71.1$ & $71.6$ & $70.7$  \\

	\dann & $76.1$ & $75.3$ & $73.7$ & $73.1$ & $72.8$ & $72.5$ & $72.6$ & $72.0$ & $74.6$ & $72.8$ & $72.8$ & $72.8$ & $73.8$ & $73.1$ \\

	\eda & $71.5$ & $70.3$ & $69.5$ & $67.7$ & $69.3$ & $68.7$ & $65.1$ & $64.6$ & $70.1$ & $68.9$ & $69.7$ & $68.0$ & $69.2$ & $68.0$  \\

	\rcrr & $75.3$ & $74.3$ & $72.8$ & $71.3$ & $72.3$ & $71.7$ & $67.4$ & $67.5$ & $72.9$ & $71.2$ & $73.0$ & $71.8$ & $72.3$ & $71.3$ \\ 

\hline

	\dcrr & $76.5$ & $75.3$ & $73.5$ & $72.4$ & $72.7$ & $72.6$ & $69.9$ & $70.3$ & $73.6$ & $72.2$ & $73.3$ & $72.3$ & $73.2$ & $72.5$ \\

	\rcdr & $75.0$ & $74.4$ & $72.5$ & $71.0$ & $72.2$ & $72.3$ & $69.9$ & $70.1$ & $72.3$ & $71.3$ & $73.2$ & $72.3$ & $72.5$ & $71.9$  \\ 
\hline

	\model & $\textbf{77.5}$ & $\textbf{76.5}$ & $\textbf{75.0}$ & $\textbf{74.0}$ & $\textbf{74.5}$ & $\textbf{74.2}$ & $\textbf{74.6}$ & $74.1$ & $\textbf{76.3}$ & $74.6$ & $\textbf{74.8}$ & $74.1$ & $\textbf{75.4}$ & $\textbf{74.6}$  \\

	\modelf & $76.9$ & $76.2$ & $74.6$ & $73.3$ & $73.7$ & $73.2$ & $\textbf{74.6}$ & $\textbf{74.6}$ & $\textbf{76.3}$ & $\textbf{74.8}$ & $74.5$ & $\textbf{74.2}$ & $75.1$ & $74.4$  \\ 

\hline
\hline

	\oracle & $80.7$ & $80.5$ & $79.6$ & $79.3$ & $78.4$ & $78.8$ & $79.8$ & $79.7$ & $80.4$ & $79.2$ & $81.0$ & $80.5$ & $80.0$ & $79.7$  \\ 
	
\hline

\end{tabular}
\end{adjustbox}
\caption{Intent prediction: F1 scores for UDA and ADA intent prediction. We report the average F1 score across five or seven target domains (UDA and ADA setups respectively).}
\label{tab:mantis_summary}
\end{table*}

\section{Results}
\label{sec:extrinsic}

Tables \ref{tab:uda_reviews} and \ref{tab:ada_reviews} present sentiment classification accuracy results for the 12 UDA and 8 ADA setups, respectively. Table \ref{tab:mantis_summary} presents the average intent prediction F1 scores for each source domain, taken across all target domains, in both UDA and ADA.

\paragraph{\domaincount{} Generation  Impact} 

For sentiment classification, our model, \modelf{}, outperforms all baseline models (\noda, \dann, \eda, and \rcrr) in 10 of 12 UDA setups and in all ADA setups, exhibiting average performance gains of $1.9\%$ and $1.3\%$  over the best performing baseline model in the UDA (\dann) and the ADA (\rcrr) setups, respectively. Moreover, \model{} without filtering, is also superior to all baselines, reaching average gains of $1.6\%$ and of $0.6\%$ across all UDA and ADA setups, respectively. 
For intent prediction, \model{} (without filtering) is the best performing model, outperforming all baselines across all setups, and reaching average gains of $1.6\%$ and $1.5\%$ across all UDA and ADA setups, respectively. Since many intent examples are not domain-specific, our filtering mechanism tends to easily remove their \model{} generated \domaincount s. We believe that this is the reason for the small degradation in \modelf{} performance compared to \model{}. However, \modelf{} still consistently outperforms all baselines. 
These results highlight the impact of \domaincount{} generation on model robustness in low-resource setups. Finally, our models are also stable: Their std is lower than all baselines (see \S\ref{sec:std}).


\paragraph{Ablation Models}
The tables further demonstrate that \modelf{} outperforms its ablation models (\S~\ref{sub:models-and-baselines}), namely \dcrr{} and \rcdr{}, in 10 of 12 and 7 of 8 UDA and ADA sentiment classification setups, respectively, and the same holds for \model{} across all intent prediction setups. Furthermore, in sentiment classification, \modelf{} achieves an average error reduction of $11.2\%$ and $5.0\%$ in UDA and ADA, respectively, over the strongest ablation model (\rcdr), while in intent prediction \model{} achieves a reduction of $8\%$ and $7.6\%$, in both setups, respectively. 
Finally, our results demonstrate the importance of inappropriate \domaincount s disqualification, as in the task of sentiment classification, \modelf{} outperforms \model{} in 8 of 12 UDA setups and in all ADA setups. On the other hand, when non domain-specific examples are frequent, filtering might lead to small performance degradation, as happens in the intent prediction task.
Our results hence stress the importance of each of \model 's algorithmic components, i.e. \textit{domain-corruption} (\S~\ref{sub:domain-coruption} \modelf{} vs \rcdr ) and \textit{oriented-reconstruction} (\S~\ref{sub:oriented-reconstruction} \modelf{} vs \dcrr ).
   


\paragraph{Complementary Effect with SOTA Models}

We notice that \modelf{} replicates the average performance of \perl{} \citep{ben2020perl}, the UDA SOTA, in sentiment classification. However, since \perl{} is based on a different architecture than the rest of the models (BERT vs T5), the models are not directly comparable. \perl{} is a pivot-based representation learning method for DA, which applies pre-training on unlabeled target data and is hence relevant only for UDA. Since \model{} implements a different approach to DA (\domaincount{} generation), we check for the complementary effect of these models: \perlaug{} first augments the labeled data with \domaincount s and then continues with the \perl{} pipeline.
As reported in Table~\ref{tab:uda_reviews}, \perlaug{} outperforms \perl{} in 8 of 12 UDA setups,  providing an average improvement of $0.7\%$. Furthermore, the average std of \perlaug{} is $2.1$ compared to $3.6$ of \perl{} (\S\ref{sec:std}). This stresses the stability of \perlaug{} across these challenging setup \citep{DBLP:conf/acl/ZiserR19}.

 Unfortunately, we cannot perform an equivalent comparison in the ADA setup, since its SOTA models \citep{ben2021pada, DBLP:conf/emnlp/WrightA20a} employ labeled data from multiple sources. Likewise, since \perl{} is not designed for multi-label prediction, we could not apply it to intent prediction.
To the best of our knowledge, we are the first to effectively perform single-source ADA. 

\paragraph{Training Size Effect}
We would next like to understand the effect of \domaincount s generated by \model{} on classifiers trained with manually labeled training sets of various sizes. 
Figure~\ref{fig:labeled_size} shows that the effect of \domaincount{} augmentation vanishes when the unaugmented classifier reaches accuracy above $85\%$ and a performance plateau (visualized as an elbow in the curve).  
These results support our hypotheses that low-resource DA scenarios may result in a model that latch on spurious domain correlations, impeding its performance. Accordingly, generating \domaincount s by intervening on the domain essentially reduces the reliance on domain-specific information and spurious correlations.

\section{Conclusions}
\label{sec:conclusions}

We presented \model{}, a corrupt-and-reconstruct approach for generating domain-counterfactuals (\domaincount s) and apply it as a data augmentation method in low-resource DA. We hypothesized that \domaincount s may mitigate the reliance on domain-specific features and on spurious correlations and help generalize out of domain. 

Our augmentation strategy yields robust models that outperform strong baselines across many low-resource DA setups. In future work we would like to further improve the controllable generation quality of \model, potentially extending it to control for multiple attributes. Moreover, we would like our methodology to address additional NLP tasks and DA setups.

\section*{Acknowledgements}
We would like to thank the action editor and the reviewers, as well as the members of the IE@Technion NLP group for their valuable feedback and advice. This research was partially funded by an ISF personal grant No. 1625/18.

\bibliographystyle{acl_natbib}
\bibliography{acl}

\clearpage
\appendix
\section{Additional Generated Examples}
\label{sec:examples}

We provide additional \domaincount{}s generated by \model{} and examples generated by \vae . 

\paragraph{(1) \domaincount s generated by \model, for different source-target domain pairs:} \mbox{}

\noindent\underline{(1.1) Original, \textbf{DVD}}:
\\ the best show on t.v. Somehow Denis Leary and Peter Tolan managed to improve on an already amazing show. Rescue me continues to be the best show out there. \smallskip

\noindent\underline{\textbf{DVD $\rightarrow$ Airline}}: 
\\ the best service on board. Somehow staff and flight crew were very good and Tolan managed to improve on an already amazing flight. iberia for me continues to be the best airline out there. \smallskip

\noindent\underline{\textbf{DVD $\rightarrow$ Electronics}}: \\ the best wireless router out there. Somehow ipaq and ethernet tolan managed to improve on an already amazing router. ethernet to me continues to be the best wireless out there. \smallskip

\noindent\underline{\textbf{DVD $\rightarrow$ Kitchen}}: 
\\ the best oven on the market. Somehow a little bit of work and chef Tolan managed to improve on an already amazing oven. this for me continues to be the best oven out there. \\




\noindent\underline{(1.2) Original, \textbf{Airline}}: 
\\ Took flight on a one way basis as return dates were subject to change. I am a top tier AA flier and the addition to Air Berlin gives some good connection options. Aircraft was a new Airbus with standard seat pitch and seat comfort in economy was uncomfortable. The IFE is good with each seat having independent controls and a reasonable selection of movies etc. Unfortunately the selection is via a touch screen buried in your seat. \smallskip

\noindent\underline{\textbf{Airline $\rightarrow$ DVD}}: 
\\ Took this movie on a trial basis as release dates are subject to change. I am a top ten fan and the addition to the bonus disc gives some good insight. The storyline with standard actors and dialogue in the main role was well done. The plot is well done with each actor having independent controls and a reasonable script. Unfortunately the movie is not about a man who is buried in your soul. \smallskip

\noindent\underline{\textbf{Airline $\rightarrow$ Electronics}}: 
\\ Took this item on a trial basis as release dates are subject to change. I am a top ten speaker and the addition to the base gives some great options. I have an ipod with standard earbuds and the built in microphone was very nice. The sound is clear with each speaker having independent controls and a reasonable bass response. Unfortunately the remote is not a good buy buried in your pocket. \smallskip

\noindent\underline{\textbf{Airline $\rightarrow$ Kitchen}}: \\ Took this item on a regular basis as delivery dates are subject to change. I am a top cook and the addition to the nonstick gives some good heat retention. I use it with standard pans and the clean up in the dishwasher was easy. The pan is very nice and each pan having independent controls and a reasonable size. Unfortunately the lid is too small for a stovetop and buried in your cupboard. \\

 \noindent\underline{(1.3) Original, \textbf{Electronics}}:
\\ "How I can test it?" perhaps you don't even need one? build a prototype circuit and test the current arrangement, and see what the noise is like. For a supposedly non-moving reference, see what the ADC is reading like. If it stays steady, and noise is less than 1 LSB, then don't do anything.
 
\noindent\underline{\textbf{Electronics $\rightarrow$ askubuntu}}: \\ "How I can test it?" perhaps you don't even need one? remount your hard drive and test the sda arrangement, and see what the sda is like. For a supposedly non-moving reference, see what the sda is reading like. If it stays in memory, and the sda is less than 1 LSB, then don't do anything. \smallskip
 
\noindent\underline{\textbf{Electronics $\rightarrow$ DBA}}: \\ "How I can test it?" perhaps you don't even need one? build a table and test the column arrangement, and see what the index is like. For a supposedly non-moving reference, see what the index is reading like. If it stays stale, and the load is less than 1 byte, then don't do anything. \smallskip
 
\noindent\underline{\textbf{Electronics $\rightarrow$ Physics}}: \\ "How I can test it?" perhaps you don't even need one? heat it up and test the particle arrangement, and see what the velocity is like. For a supposedly non-moving reference, see what the waveform is reading like. If it stays steady, and the speed is less than 1 m/s, then don't do anything. \smallskip
 
 \noindent\underline{\textbf{Electronics $\rightarrow$ Statistics}}: \\ "How I can test it?" perhaps you don't even need one? build a regression model and test the sp arrangement, and see what the plot is like. For a supposedly non-moving reference, see what the plot is reading like. If it stays constant, and the mean is less than 1\%, then don't do anything. \\

\noindent\underline{(1.4) Original, \textbf{Statistics}}: 
\\  While thinking about methods on how to predict a dependent variable that is ordinal and constrained (0, 1, 2, 3, 5 are the only possible values), I found that an ordered logistic regression is an common approach used in the community and as a strategy, that make sense to me. However, I was trying to understand why another method that came into my mind is a bad idea. \smallskip

\noindent\underline{\textbf{Statistics $\rightarrow$ Apple}}: While thinking about methods on how to store music that is ordered by album and song (0, 1, 2, 3, 5 are the only possible ids), I found that an ordered playlist is an option used in the community and as a backup, that make sense to me. However, I was trying to understand why another method that came into my mind is a bad idea. \smallskip

\noindent\underline{\textbf{Statistics $\rightarrow$ askubuntu}}: While thinking about methods on how to install a package that is compiled and installed (0, 1, 2, 3, 5 are the only possible repos), I found that an ordered deb package is an common package used in the community and as a result, that make sense to me. However, I was trying to understand why another method that came into my mind is a bad idea. \smallskip
 
\noindent\underline{\textbf{Statistics $\rightarrow$ DBA}}: While thinking about methods on how to store a dependent table that is a single and unique column (0, 1, 2, 3, 5 are the only column), I found that an ordered index is an approach used in the community and as a strategy, that make sense to me. However, I was trying to understand why another method that came into my mind is a bad idea. \smallskip
  
\noindent\underline{\textbf{Statistics $\rightarrow$ Electronics}}:  While thinking about methods on how to design a circuit that is programmble and constrained (0, 1, 2, 3, 5 are the only inputs), I found that an ordered voltage is an analog circuit used in the community and as a switch, that make sense to me.However, I was trying to understand why another method that came into my mind is a bad idea. \smallskip

 \noindent\underline{\textbf{Statistics $\rightarrow$ Physics}}: While thinking about methods on how to construct a particle that is ordered and constrained (0, 1, 2, 3, 5 are the only constraints), I found that an ordered field theory is an approach used in the community and as a general relativist, that make sense to me.However, I was trying to understand why another method that came into my mind is a bad idea. \\

\paragraph{(2) \domaincount s generated by \model{} and their masked versions:} \mbox{}

\noindent\underline{(2.1) Original, \textbf{Airline}}:
\\ One major problem was that the day I flew there was a fuel service strike so I was delayed for over 4 hours. I frequently check my flight status especially when I leave, throughout that day I had checked at least 4 times. I only found out about it when I was at the check in counter. There was no email, no automated phone call, nothing to notify me. I was stuck waiting for hours doing nothing. \smallskip

\noindent\underline{Masked text, \textbf{Airline $\rightarrow$ Kitchen}}:
\\ One major problem was that the day I [MASK] a [MASK] strike so I was [MASK] for [MASK]. I frequently [MASK] my [MASK] especially when I leave, [MASK] I had [MASK] at least 4 times. I [MASK] out about it when I [MASK]. [MASK], no [MASK] call, [MASK] me. [MASK] for [MASK] doing nothing. \smallskip

\noindent\underline{\textbf{Airline $\rightarrow$ Kitchen}}: \\ One major problem was that the day I got home there was a labor strike so I was left without a coffee maker for the night. I frequently refill my coffee especially when I leave, and I had to replace it at least 4 times. I found out about it when I got home. I sent it back to the store, no one came by to call, and they didn't help me. I sat for over an week doing nothing. 
\smallskip




\noindent\underline{(2.2) Original, \textbf{Electronics}}:
\\ disappointed with the mouse Though I like the keyboard, I've found two serious problems with the mouse. First, it is very sensitive to movement. I've tried making some adjustments but it is still difficult to control. Finally and more importantly, the mouse burns a set of AA batteries about every 8 days. I only use the mouse about 6 hours a week so I should be getting a lot more battery life. I've bought other Logitech mouse. \smallskip

\noindent\underline{Masked text, \textbf{Electronics $\rightarrow$ DVD}}:
\\ disappointed with the [MASK] Though I like the [MASK], [MASK] two serious problems with the [MASK]. First, it is very [MASK] to movement. [MASK] making some [MASK] but it is still difficult [MASK]. Finally and more importantly, the [MASK] about every 8 days. I [MASK] the [MASK] about 6 [MASK] week so I should be getting a lot more [MASK] life. I've [MASK] other [MASK]. \smallskip

\noindent\underline{\textbf{Electronics $\rightarrow$ DVD}}: \\ disappointed with the workout. Though I like the workout, I have two serious problems with the workout. First, it is very slow to movement. I'm making some progress but it is still difficult to follow. Finally and more importantly, the workout only goes on about every 8 days. I do the workout about 6 days a week so I should be getting a lot more exercise in my life. I've seen other workouts that aren't slow. 
  \medskip



\noindent\underline{(2.3) Original, \textbf{Physics}}:
\\  @PaulStonaha: elastic forces cannot provide extra pressure in excess of 10\% (balloons pop at such excess pressure), so elastic properties of balloons should not have significant effect.
 \smallskip

\noindent\underline{Masked text, \textbf{Physics $\rightarrow$ askubuntu}}:
\\  @PaulStonaha: [MASK] cannot provide extra [MASK] in excess of 10\% ([MASK] pop [MASK] excess [MASK]), so [MASK] should not have significant effect.
 \smallskip

\noindent\underline{\textbf{Physics $\rightarrow$ askubuntu}}:
\\  @PaulStonaha: dpkg cannot provide extra partitions in the in excess of 10\% (ubuntu pop installs excess partitions), so redownload should not have significant effect. 

\paragraph{(3) \domaincount s generated by \model{} while providing different orientation vectors:} \mbox{}

\noindent\underline{(3.1) Original, \textbf{Airline}}:
\\ It was a fantastic flight crew helpful and smiling. All announcements very clear understandable and most important things for me multi - language Turkish English Russian and Kazakh. The way aircraft was new. \smallskip

\noindent\underline{\textbf{Airline $\rightarrow$ DVD}, Orientation: ``dvd''}:
\\ It was a fantastic movie. All the main actors were very clear understandable and most important things for me multi - language Russian and Kazakh. The way they talk. \smallskip

\noindent\underline{\textbf{Airline $\rightarrow$ DVD}, Orientation: ``character''}:
\\ It was a fantastic movie. All the main character's speech was very clear understandable and most important things for me multi - language Russian and Kazakh. The way it was done was great. \smallskip

\noindent\underline{\textbf{Airline $\rightarrow$ DVD}, Orientation: ``actor''}:
\\ It was a fantastic movie. All actors very clear understandable and most important things for me multi - language Russian and Kazakh. The way they act. \smallskip

\noindent\underline{\textbf{Airline $\rightarrow$ DVD}, Orientation: ``plot''}:
\\ It was a fantastic movie. All the plots were very clear understandable and most important things for me multi - language Russian and Kazakh. The way it was told was very good. \\

\paragraph{(4) Examples generated by the \vae{} baseline, controlling both for the domain and the label:} \mbox{}

\noindent\underline{\textbf{Airline}, \textbf{Positive}}:
\\ i paid for a trip to hawaii. i was traveling with aeroflot. i was pleasantly surprised. i was pleasantly surprised. the service was excellent and service. i paid for \$50 for business class. i am very satisfied with this airline. \smallskip

\noindent\underline{\textbf{Airline}, \textbf{Positive}}:
\\ one of the most memorable movie ever made. i think this movie is a silly comedy, but i was a little silly "attitude of" the "buddy" . "attitude" attitude of the robots, but i was a little silly job of the movie. 
 
\noindent\underline{\textbf{Electronics}, \textbf{Negative}}: \\ not worth the money for my ipod nano. i bought this product for my 3 year old and i am not sure why i am not sure why i am not sure why i am not disappointed. \smallskip

\noindent\underline{\textbf{Kitchen}, \textbf{Positive}}:
\\ broken broken after a broken set of my mother and i needed a gift for my sister. i was skeptical about how to do it. i was able to use it to my dishwasher safe and i was delighted with a silverware. i would recommend it

\section{Implementation Details}
\label{sub:implementation}

\subsection{URLs of Code and Data}
\label{sub:urls}

\begin{itemize}
  \item \textbf{\model{} Repository} - code and datasets: \url{https://github.com/nitaytech/DoCoGen}
  \item \textbf{HuggingFace} \citep{wolf2020transformers} - code and pretrained weights for the T5 model and tokenizer: \url{https://huggingface.co/}
  \item \textbf{SentenceTransformers} \citep{reimers2019sentence} - code and pretrained weights of a LM. We use this LM to extract the embeddings of input examples, and then calculate the cosine similarity between them to match examples in the \oracle{} model: \url{https://www.sbert.net/}
  \item \textbf{PERL} \cite{ben2020perl} - A SOTA unsupervised domain adaptation model: \url{https://github.com/eyalbd2/PERL}
  \item \textbf{EDA} \citep{wei2019eda} - \url{https://github.com/jasonwei20/eda_nlp}
  \item \textbf{VAE} - based on the controllable generation model of \citet{russo2020control}: \url{https://github.com/DS3Lab/control-generate-augment}
\end{itemize}

\subsection{Hyperparameters and Setups}
\label{sub:hyperparameters}

\paragraph{Data Preprocessing}

We truncate each example to 96 tokens, using the HuggingFace T5-base tokenizer. The hyper-parameter was set to 96 due to computation reasons and since the median number of words in the labeled examples was 89. When an example is longer than 96 tokens, we keep the first 96 tokens. For examples from the Airline domain, before truncating, we remove the first sentence since it mostly contains details about the flight (like ``from JPK to LAX'').

\paragraph{\model} 

\underline{Masking:} We estimate $P(\mathcal{D}|w)$ for uni-grams, bi-grams and tri-grams which appear in the unlabeled data in at least 10 examples. We use the NLTK Snowball stemmer to stem each word of the n-grams. The smoothing hyperparameters in the computation of $P(\mathcal{D}|w)$ are set to be $1, 5$ and $7$ for uni-grams, bi-grams and tri-grams, respectively. We use a $\tau=0.08$ threshold and mask additional 5\% of the training examples (in order to add noise between training epochs). We set $\tau=0.08$ since it resulted in the successfully domain alternation of more than 80\% examples. For \rcrr{} and \rcdr{} we randomly mask 15\%  of the examples (the standard ratio for MLM). 

\underline{Controllable Model:} 
We use $K=4$ orientation vectors for each unlabeled domain and initialize them with the following representing words for the sentiment dataset: Airline: \{airline, flight, seat, staff\}, DVD: \{dvd, character, actor, plot\}, Electronics: \{electronics, ipod, router, software\}, Kitchen: \{kitchen, dishwasher, pan, oven\} and for the MANtIS dataset: Apple: \{apple, itunes, iphone, nacbook\}, askubuntu: \{askubuntu, ubuntu, apt, deb\}, DBA: \{dba, database, sql, query\}, Electronics: \{electronics, schematic, voltage, circuit\}, Physics: \{physics, gravity, particle, quantom\}, Statistics: \{stats, regression, logists, variance\}. 

The controllable model is based on a pretrained HuggingFace T5-base model. We train it on the unlabeled data for 20 epochs and pick the model
whose generated examples for an unlabeled held-out set are of the highest domain-accuracy (\domainrel ).\footnote{The domain accuracy is measured by a domain-classifier trained on the unlabeled data and that is based on the T5 encoder architecture.} Training is performed with the AdamW optimizer \citep{loshchilov2018decoupled} with a learning rate parameter of 5e-5 and a weight decay parameter of 1e-5. For \rcrr{} and \rcdr{} we pick the best models based on a MLM loss computed on a held-out set. In the example generation step we use a Beam Search decoding method with a beam size of 4.

\paragraph{\vae}
As described in the main paper, our \vae{} implementation is based on \citet{russo2020control}. To adjust the model for the purposes of this research, we control the task label and the domain label of each generated review. We train the model on the entire labeled data and unlabeled data that is available from four domains: A, D, E, and K, for a total of $8000$ labeled reviews and $104075$ unlabeled reviews. We train the \vae{} for $60$ epochs, concatenating sentences with more than $96$ tokens, and applying a batch size of $32$. The rest of the hyperparameters were set to the values described in  \citet{russo2020control}.

\paragraph{DA Evaluation}

\underline{Data Augmentation}
Given a labeled example from the source domain, we generate $K \cdot N=16$ examples by \model , where $K$ is the number of orientation vectors of each  domain and $N$ is the number of unlabeled domains. We use the generated examples for data augmentation for the task classifiers. For all augmentation models, we apply an augmentation ratio identical to the one used for \model, yielding augmented training sets of the same size. For \noda{} and \dann{} we duplicate the training set $K\cdot N$ times, thus the number of training steps of all the classifiers is identical. For EDA we use the default hyperparameters.  

\underline{Task Classifiers} All classifiers are based on the T5-encoder architecture equipped with a linear layer, except from \perl{} which is based on the BERT architecture. We train the classifiers for 5 epochs with a batch size of 64 and pick the best model based on the performance on the validation set. Training is performed using the AdamW optimizer with learning rate parameters of 5e-5 for the encoder blocks and of 5e-4 for the linear layer. 

For the results reported in Tables~\ref{tab:uda_reviews}, \ref{tab:ada_reviews}, \ref{tab:std_uda_reviews}, \ref{tab:std_ada_reviews}, \ref{tab:mantis_summary}, \ref{tab:mantis_uda} and \ref{tab:mantis_ada} we employ a training set that consists of 100 examples and a validation set with 25 examples. To increase the robustness of the results in our small labeled training set setup, we train 25 classifiers, each using a different randomized seed and a randomly sampled training set. We report the average performance of these classifiers on the test set.
For the results reported in Figure~\ref{fig:labeled_size}, the validation set size is 25\% of the training size. We train the classifiers on 25 different seeds and partitions for training sizes 25, 50 and 100, and 10 seeds and partitions for sizes 250, 500 and 1000.

\begin{table}
\centering
\begin{adjustbox}{width=0.48\textwidth}

\begin{tabular}{ | l || c | c | c | c | }
\hline
\multicolumn{5}{|c|}{Sentiment Classification} \\
\hline
Domain & Unlabeled & Train & Dev & Test  \\
\hline
Airline (A) & 39454 & 1700 (100) & 300 (25) & 2000 \\
\hline
DVDs (D) & 34742 & 1700 (100) & 300 (25) & 2000 \\
\hline
Electronics (E) & 13154 & 1700 (100) & 300 (25) & 2000 \\
\hline
Kitchen (K) & 16786 & 1700 (100) & 300 (25) & 2000 \\
\hline
Books (B) & 6001 (0) & 1700 (0) & 300 (0) & 2000 \\
\hline
Restaurant (R) & 25000 (0) & 1700 (0) & 300 (0) & 2000 \\
\hline
\multicolumn{5}{|c|}{Intent Classification} \\
\hline
Domain & Unlabeled & Train & Dev & Test  \\
\hline
Apple (AP) & 24752 & 354 (100) & 142 (25) & 196 \\
\hline
DBA (DB) & 25121 & 311 (100) & 138 (25) & 199 \\
\hline
Electronics (EL) & 27192 & 664 (100) & 276 (25) & 397 \\
\hline
Physics (PH) & 25675 & 142 (100) & 68 (25) & 78 \\
\hline
Statistics (ST) & 25743 & 176 (100) & 72 (25) & 102 \\
\hline
Askubuntu (UB) & 26930 & 1096 (100) & 418 (25) & 610 \\
\hline
DIY (DI) & 7383 (0) & 0 (0) & 0 (25) & 180 \\
\hline
English (EN) & 14734 (0) & 0 (0) & 0 (0) & 189 \\
\hline
Gaming (GA) & 14050 (0) & 0 (0) & 0 (0) & 117 \\
\hline
GIS (GI) & 25291 (0) & 0 (0) & 0 (0) & 418 \\
\hline
Sci-Fi (SC) & 10145 (0) & 0 (0) & 0 (0) & 109 \\
\hline
Security (SE) & 18302 (0) & 0 (0) & 0 (0) & 109 \\
\hline
Travel (TR) & 6687 (0) & 0 (0) & 0 (0) & 61 \\
\hline
Worldbuilding (WO) & 6044 (0) & 0 (0) & 0 (0) & 54 \\
\hline
\end{tabular}
\end{adjustbox}
\caption{Number of available samples in each domain. Numbers in parenthesis represent the amount of samples used for each DA setup.}
\label{tab:datasets}
\end{table}

\subsection{Masking}
\label{sub:masking}
Table~\ref{tab:masking-per} presents the average percentage of masked tokens in the corruption step of \model{} (see \S\ref{sub:domain-coruption}), in the sentiment classification dataset. Overall, the average percentage of masked tokens in a single review is $25.2$. These statistics emphasize the large gap between original reviews and their \domaincount s.  

\begin{table}
\small
\centering
\begin{tabular}{ | c | c | c | c | c |}

\hline
 \textbf{$\nearrow$} & \textbf{A}  & \textbf{D}  & \textbf{E} & \textbf{K}  \\
\hline 
	\textbf{A} & $15.2$ & $37.9$ & $37.3$ & $38.0$  \\ 

\hline 

	\textbf{D} & $25.0$ & $16.5$ & $24.0$ & $23.9$   \\ 

\hline 

    \textbf{E} & $27.8$ & $26.7$ & $15.7$ & $19.7$  \\ 

\hline 

    \textbf{K} & $30.2$ & $28.0$ & $21.1$ & $15.7$  \\ 

\hline

\end{tabular}
\caption{Percentage of tokens of the original examples that were masked by \model{} in the sentiment classification dataset. The left column indicates the source domain and the top row indicates the target domain.}
\label{tab:masking-per}
\end{table}


\begin{table*}
\centering
\begin{adjustbox}{width=\textwidth}

\begin{tabular}{ | l || c | c | c | c | c | c |c | c | c | c | c | c || c | }

\hline
  & \textbf{A $\rightarrow$ D} & \textbf{A $\rightarrow$ E} & \textbf{A $\rightarrow$ K} & \textbf{D $\rightarrow$ A} & \textbf{D $\rightarrow$ E} & \textbf{D $\rightarrow$ K} & \textbf{E $\rightarrow$ A} & \textbf{E $\rightarrow$ D} & \textbf{E $\rightarrow$ K} & \textbf{K $\rightarrow$ A} & \textbf{K $\rightarrow$ D} & \textbf{K $\rightarrow$ E} & \textbf{AVG} \\
  \hline
  
	\noda & $7.8$ & $6.0$ & $6.8$ & $6.7$ & $5.7$ & $5.4$ & $2.6$ & $4.7$ & $3.0$ & $6.8$ & $4.1$ & $2.9$ & $5.2$ \\

	\dann & $5.4$ & $4.9$ & $5.8$ & $\textbf{5.2}$ & $4.5$ & $4.4$ & $3.1$ & $3.4$ & $3.4$ & $2.8$ & $4.4$ & $2.5$ & $4.1$ \\

	\eda & $6.1$ & $5.7$ & $5.8$ & $7.1$ & $6.8$ & $5.4$ & $4.4$ & $4.9$ & $3.5$ & $6.1$ & $4.5$ & $2.9$ & $5.3$ \\

	\rcrr & $6.8$ & $4.9$ & $5.2$ & $5.7$ & $5.1$ & $4.7$ & $3.2$ & $4.3$ & $2.8$ & $5.5$ & $5.1$ & $3.3$ & $4.7$ \\ 

\hline

	\dcrr & $8.0$ & $6.8$ & $7.5$ & $6.8$ & $6.1$ & $5.3$ & $3.0$ & $3.1$ & $2.0$ & $5.0$ & $4.8$ & $3.1$ & $5.1$ \\

	\rcdr & $7.6$ & $4.9$ & $5.4$ & $6.7$ & $5.6$ & $4.7$ & $3.8$ & $\textbf{2.0}$ & $2.0$ & $7.4$ & $4.8$ & $3.1$ & $4.8$ \\ 

\hline

	\model & $5.9$ & $4.7$ & $5.1$ & $5.5$ & $4.0$ & $3.5$ & $\textbf{1.9}$ & $2.5$ & $2.3$ & $2.2$ & $2.9$ & $1.9$ & $3.5$ \\

	\modelf & $\textbf{4.9}$ & $\textbf{4.3}$ & $\textbf{4.8}$ & $\textbf{5.2}$ & $\textbf{3.8}$ & $\textbf{3.1}$ & $2.0$ & $2.3$ & $\textbf{1.9}$ & $\textbf{2.1}$ & $\textbf{2.0}$ & $\textbf{1.7}$ & $\textbf{3.2}$ \\ 
  \hline
  \hline
  
  \perl & $8.3$ & $5.4$ & $4.6$ & $\underline{2.0}$ & $6.3$ & $\underline{1.2}$ & $2.3$ & $2.1$ & $\underline{0.7}$ & $4.7$ & $4.1$ & $1.4$ & $3.6$ \\
  
  \perlaug & $\underline{2.2}$ & $\underline{0.9}$ & $\underline{2.7}$ & $3.0$ & $\underline{1.6}$ & $2.1$ & $\underline{1.9}$ & $\underline{1.0}$ & $2.8$ & $\underline{4.1}$ & $\underline{1.7}$ & $\underline{0.9}$ & $\underline{2.1}$ \\

    \hline
    \hline

	\oracle & $1.6$ & $1.2$ & $1.7$ & $1.8$ & $1.0$ & $1.4$ & $0.8$ & $1.2$ & $1.0$ & $1.4$ & $2.9$ & $0.9$ & $1.4$ \\

\hline

\end{tabular}
\end{adjustbox}
\caption{Sentiment classification: Standard deviations for each source and target domain pair in the UDA setup. \textbf{Bold} numbers mark the best performing T5-based model, and \underline{underlined} numbers mark the best performing \perl{} model.}
\label{tab:std_uda_reviews}
\end{table*}

\begin{table*}
\centering
\begin{adjustbox}{width=0.8\textwidth}

\begin{tabular}{ | l || c | c | c | c | c | c | c | c || c | }

\hline
  & \textbf{A $\rightarrow$ B} & \textbf{A $\rightarrow$ R} & \textbf{D $\rightarrow$ B} & \textbf{D $\rightarrow$ R} & \textbf{E $\rightarrow$ B} & \textbf{E $\rightarrow$ R} & \textbf{K $\rightarrow$ B} & \textbf{K $\rightarrow$ R} & \textbf{AVG} \\
  \hline
  
	\noda & $8.0$ & $6.3$ & $3.5$ & $3.7$ & $5.7$ & $4.0$ & $4.1$ & $2.7$ & $4.8$ \\

	\dann & $6.5$ & $6.2$ & $3.3$ & $3.7$ & $3.3$ & $2.2$ & $3.5$ & $4.2$ & $4.1$ \\

	\eda & $\textbf{5.9}$ & $4.9$ & $4.1$ & $5.0$ & $5.2$ & $4.3$ & $5.0$ & $3.5$ & $4.7$ \\

	\rcrr & $7.0$ & $4.8$ & $2.9$ & $3.5$ & $5.2$ & $2.9$ & $3.5$ & $2.4$ & $4.0$ \\ 

\hline

	\dcrr & $8.2$ & $6.2$ & $2.8$ & $4.0$ & $3.7$ & $\textbf{1.6}$ & $4.4$ & $3.1$ & $4.2$ \\

	\rcdr & $7.8$ & $4.9$ & $2.9$ & $4.6$ & $\textbf{2.6}$ & $1.9$ & $3.4$ & $3.3$ & $3.9$ \\ 

\hline

	\model & $7.0$ & $5.7$ & $2.4$ & $3.4$ & $3.2$ & $\textbf{1.6}$ & $2.6$ & $2.4$ & $3.5$ \\

	\modelf & $6.0$ & $\textbf{4.0}$ & $\textbf{2.0}$ & $\textbf{3.3}$ & $3.0$ & $1.7$ & $\textbf{1.9}$ & $\textbf{1.3}$ & $\textbf{2.9}$ \\ 

\hline
\hline

	\oracle & $2.1$ & $2.3$ & $2.0$ & $1.6$ & $1.6$ & $1.8$ & $2.4$ & $1.4$ & $1.9$ \\ 

\hline

\end{tabular}
\end{adjustbox}
\caption{Sentiment classification: Standard deviations for each source and target domain pair in the ADA setup.}
\label{tab:std_ada_reviews}
\end{table*}

\section{Ablation Results}

\subsection{Standard Deviations}
\label{sec:std}

Each of the numbers reported in the main result tables of the main paper is the average of 25 repetitions, across seeds and training sets. We hence also report here the standard deviations of these results, which indicate on the stability of the participating models.

The standard deviations for the UDA and ADA setups of sentiment classification are presented in Tables~\ref{tab:std_uda_reviews} and \ref{tab:std_ada_reviews}, respectively. \modelf{} outperforms all baseline models (\noda, \dann, \eda, and \rcrr) in 11 of 12 UDA setups and in 6 of 8 ADA setups, demonstrating a lower average standard deviation and an improvement of $22.0\%$ and $27.5\%$ in the UDA and the ADA setups, respectively, over the best performing baseline model. Moreover, \model{} without filtering is also superior to all baselines. These results highlight the impact of \domaincount{} generation on model stability in low-resource DA setups. 

As noted in the main paper, we also evaluate the complementary effect of \model{} and \perl{}, a SOTA model for UDA. Tables~\ref{tab:std_uda_reviews} shows that \perlaug{} achieves the lowest average standard deviation, improving \perl{} by $42\%$.  \perlaug{}  is hence the best performing model both in terms of accuracy (see main paper) and in terms of standard deviation (stability).

Tables~\ref{tab:mantis_uda} and \ref{tab:mantis_ada} report the F1 scores and the standard deviations for the UDA and ADA setups of intent classification, respectively. As in the case of sentiment classification, \modelf{} and \model{} are superior to all baselines, achieving lower standard deviation results in the majority of setups. The tables provide additional information regarding the F1 results presented in the main paper (Table~\ref{tab:mantis_summary}), reporting F1 scores obtained for each source/target pair experiment.

\begin{table*}
\centering
\begin{adjustbox}{width=\textwidth}

\begin{tabular}{ | l || c | c | c | c | c | c | c | c | c | c |}

\hline
& \textbf{AP $\rightarrow$ DB} & \textbf{AP $\rightarrow$ EL} & \textbf{AP $\rightarrow$ PH} & \textbf{AP $\rightarrow$ ST} & \textbf{AP $\rightarrow$ UB} & \textbf{DB $\rightarrow$ AP} & \textbf{DB $\rightarrow$ EL} & \textbf{DB $\rightarrow$ PH} & \textbf{DB $\rightarrow$ ST} & \textbf{DB $\rightarrow$ UB} \\

\hline
	\noda & $77.2 \pm 5.3$ & $76.8 \pm 5.2$ & $71.4 \pm 9.3$ & $74.6 \pm 6.7$ & $77.3 \pm 4.3$ & $74.6 \pm 6.3$ & $74.1 \pm 6.1$ & $66.4 \pm 11.6$ & $72.6 \pm 7.8$ & $73.2 \pm 5.3$ \\

	\dann & $77.7 \pm 5.0$ & $77.0 \pm 4.9$ & $73.2 \pm 9.3$ & $74.4 \pm 6.5$ & $78.0 \pm 4.2$ & $76.1 \pm 5.9$ & $74.8 \pm 5.5$ & $69.8 \pm \textbf{9.7}$ & $73.4 \pm 7.0$ & $74.7 \pm 5.1$ \\

	\eda & $73.7 \pm 5.6$ & $72.4 \pm 6.1$ & $65.8 \pm 9.3$ & $71.1 \pm 6.2$ & $74.4 \pm 4.8$ & $71.3 \pm 6.6$ & $70.6 \pm 6.9$ & $63.9 \pm 10.2$ & $69.5 \pm 7.8$ & $72.0 \pm 5.2$ \\

	\rcrr & $77.2 \pm 5.7$ & $76.8 \pm 4.5$ & $70.9 \pm 9.1$ & $74.2 \pm 6.3$ & $77.4 \pm 4.5$ & $75.1 \pm 5.6$ & $75.1 \pm 5.3$ & $66.3 \pm 10.2$ & $73.5 \pm 7.6$ & $73.8 \pm 5.1$ \\ 
\hline

	\dcrr & $78.4 \pm \textbf{4.4}$ & $77.7 \pm 4.8$ & $72.4 \pm 8.4$ & $75.9 \pm 5.7$ & $78.3 \pm 4.2$ & $75.8 \pm 5.7$ & $75.2 \pm 4.9$ & $68.3 \pm 10.4$ & $73.4 \pm \textbf{6.4}$ & $74.7 \pm 4.6$ \\

	\rcdr & $77.8 \pm 5.2$ & $76.6 \pm 5.0$ & $69.5 \pm 9.2$ & $74.1 \pm 6.5$ & $77.2 \pm 4.2$ & $75.4 \pm 6.1$ & $74.0 \pm 6.3$ & $66.4 \pm \textbf{9.7}$ & $72.6 \pm 8.2$ & $74.3 \pm 5.2$ \\ 

\hline

	\model & $\textbf{79.2} \pm \textbf{4.4}$ & $\textbf{78.6} \pm 3.9$ & $\textbf{74.0} \pm \textbf{7.6}$ & $\textbf{76.7} \pm 5.3$ & $\textbf{78.9} \pm \textbf{3.7}$ & $\textbf{77.1} \pm \textbf{5.1}$ & $\textbf{76.3} \pm \textbf{4.8}$ & $\textbf{70.7} \pm \textbf{9.7}$ & $\textbf{74.9} \pm 6.5$ & $\textbf{75.9} \pm \textbf{4.1}$ \\

	\modelf & $78.6 \pm 4.6$ & $78.2 \pm \textbf{3.8}$ & $73.1 \pm \textbf{7.6}$ & $76.0 \pm \textbf{4.7}$ & $78.6 \pm \textbf{3.7}$ & $77.0 \pm 6.3$ & $75.8 \pm 5.2$ & $70.2 \pm 10.3$ & $74.4 \pm 7.3$ & $75.5 \pm 4.9$ \\ 

\hline

	\oracle & $82.6 \pm 3.2$ & $81.3 \pm 2.5$ & $79.0 \pm 5.0$ & $78.4 \pm 3.9$ & $82.3 \pm 2.4$ & $81.7 \pm 3.6$ & $80.2 \pm 3.4$ & $76.6 \pm 6.5$ & $79.5 \pm 4.6$ & $80.2 \pm 3.1$ \\

\hline
\hline

& \textbf{EL $\rightarrow$ AP} & \textbf{EL $\rightarrow$ DB} & \textbf{EL $\rightarrow$ PH} & \textbf{EL $\rightarrow$ ST} & \textbf{EL $\rightarrow$ UB} & \textbf{PH $\rightarrow$ AP} & \textbf{PH $\rightarrow$ DB} & \textbf{PH $\rightarrow$ EL} & \textbf{PH $\rightarrow$ ST} & \textbf{PH $\rightarrow$ UB} \\
\hline
	\noda & $72.8 \pm 7.2$ & $72.4 \pm 7.1$ & $67.6 \pm 9.8$ & $71.7 \pm 8.0$ & $71.3 \pm 7.4$ & $64.5 \pm 9.9$ & $67.5 \pm 9.3$ & $69.5 \pm 7.0$ & $72.8 \pm 7.7$ & $61.3 \pm 7.7$ \\

	\dann & $74.7 \pm 6.2$ & $73.7 \pm 6.5$ & $69.6 \pm 9.1$ & $72.2 \pm 7.0$ & $73.7 \pm 6.0$ & $73.1 \pm 7.2$ & $72.7 \pm 6.5$ & $73.0 \pm 5.4$ & $73.9 \pm 7.0$ & $70.4 \pm 6.2$ \\

	\eda & $70.3 \pm 6.6$ & $70.5 \pm 6.6$ & $66.1 \pm 9.3$ & $70.0 \pm \textbf{6.2}$ & $69.5 \pm 6.4$ & $61.8 \pm 6.7$ & $65.5 \pm 6.7$ & $67.2 \pm 6.2$ & $71.1 \pm 6.4$ & $60.2 \pm \textbf{5.3}$ \\

	\rcrr & $74.2 \pm 6.8$ & $73.7 \pm 6.9$ & $69.0 \pm 9.3$ & $72.0 \pm 7.7$ & $72.5 \pm 7.0$ & $65.0 \pm 8.3$ & $67.8 \pm 6.9$ & $70.0 \pm 5.9$ & $73.3 \pm 6.8$ & $60.8 \pm 6.5$ \\ 
\hline

	\dcrr & $74.5 \pm 6.8$ & $74.1 \pm 6.6$ & $69.1 \pm 8.6$ & $73.2 \pm 7.1$ & $72.5 \pm 6.5$ & $68.5 \pm 9.4$ & $70.7 \pm 8.5$ & $71.5 \pm 6.2$ & $75.1 \pm 6.6$ & $63.7 \pm 7.9$ \\

	\rcdr & $74.5 \pm \textbf{6.0}$ & $73.5 \pm 6.1$ & $67.5 \pm 9.2$ & $72.7 \pm 7.7$ & $72.6 \pm 6.6$ & $67.7 \pm 8.0$ & $70.8 \pm 6.9$ & $71.9 \pm 5.9$ & $75.5 \pm 5.8$ & $63.6 \pm 6.7$ \\ 

\hline

	\model & $\textbf{76.0} \pm \textbf{6.0}$ & $\textbf{75.9} \pm 5.6$ & $\textbf{71.0} \pm 8.6$ & $\textbf{74.6} \pm 6.3$ & $\textbf{75.0} \pm 5.1$ & $75.3 \pm 6.6$ & $\textbf{74.1} \pm 6.1$ & $75.0 \pm 4.6$ & $\textbf{76.0} \pm \textbf{5.3}$ & $72.6 \pm 6.2$ \\

	\modelf & $75.5 \pm \textbf{6.0}$ & $74.8 \pm \textbf{5.4}$ & $70.6 \pm \textbf{7.5}$ & $73.6 \pm 6.7$ & $74.2 \pm \textbf{5.0}$ & $\textbf{75.5} \pm \textbf{6.0}$ & $73.8 \pm \textbf{5.2}$ & $\textbf{75.3} \pm \textbf{4.5}$ & $75.3 \pm 5.9$ & $\textbf{72.9} \pm 5.9$ \\ 

\hline

	\oracle & $80.4 \pm 4.0$ & $78.7 \pm 4.5$ & $75.9 \pm 6.4$ & $77.9 \pm 4.9$ & $78.9 \pm 3.2$ & $81.4 \pm 3.4$ & $79.8 \pm 3.5$ & $79.9 \pm 2.4$ & $79.0 \pm 3.7$ & $79.2 \pm 2.8$ \\

\hline
\hline

& \textbf{ST $\rightarrow$ AP} & \textbf{ST $\rightarrow$ DB} & \textbf{ST $\rightarrow$ EL} & \textbf{ST $\rightarrow$ PH} & \textbf{ST $\rightarrow$ UB} & \textbf{UB $\rightarrow$ AP} & \textbf{UB $\rightarrow$ DB} & \textbf{UB $\rightarrow$ EL} & \textbf{UB $\rightarrow$ PH} & \textbf{UB $\rightarrow$ ST} \\
\hline
	\noda & $70.6 \pm 7.1$ & $73.6 \pm 5.8$ & $75.0 \pm 4.6$ & $70.6 \pm 6.8$ & $69.3 \pm 6.3$ & $74.6 \pm 6.3$ & $73.5 \pm 6.0$ & $72.7 \pm 6.6$ & $67.0 \pm 10.0$ & $72.0 \pm 6.7$ \\

	\dann & $74.7 \pm 5.1$ & $75.5 \pm 4.4$ & $76.5 \pm 4.0$ & $72.6 \pm 6.8$ & $73.8 \pm 4.4$ & $75.4 \pm 6.4$ & $74.9 \pm 5.7$ & $72.8 \pm 6.5$ & $69.3 \pm 8.7$ & $71.8 \pm 6.6$ \\

	\eda & $68.9 \pm 7.8$ & $72.5 \pm 5.9$ & $72.4 \pm 5.8$ & $68.2 \pm 7.3$ & $68.7 \pm 6.9$ & $73.2 \pm 5.9$ & $72.4 \pm 6.2$ & $70.3 \pm 6.8$ & $63.4 \pm 10.1$ & $69.4 \pm 7.3$ \\

	\rcrr & $72.0 \pm 7.4$ & $74.6 \pm 5.3$ & $75.7 \pm 4.5$ & $71.7 \pm 6.9$ & $70.3 \pm 6.9$ & $76.2 \pm 6.0$ & $75.4 \pm 5.7$ & $73.8 \pm 5.6$ & $66.9 \pm \textbf{8.6}$ & $72.5 \pm 6.9$ \\ 
\hline

	\dcrr & $72.7 \pm 6.4$ & $74.4 \pm 5.7$ & $76.6 \pm 4.1$ & $73.5 \pm 6.7$ & $70.9 \pm 5.8$ & $76.5 \pm 5.6$ & $75.2 \pm 5.6$ & $74.0 \pm 6.3$ & $68.1 \pm 9.4$ & $72.7 \pm 6.6$ \\

	\rcdr & $71.4 \pm 7.2$ & $74.1 \pm 6.0$ & $75.2 \pm 4.7$ & $70.9 \pm 6.7$ & $69.9 \pm 6.7$ & $76.6 \pm 5.5$ & $75.4 \pm \textbf{5.3}$ & $74.3 \pm 5.6$ & $66.9 \pm 9.8$ & $72.5 \pm 8.4$ \\ 

\hline

	\model & $76.4 \pm 4.6$ & $\textbf{76.4} \pm 4.5$ & $78.2 \pm 3.6$ & $\textbf{75.0} \pm 6.2$ & $75.7 \pm \textbf{3.8}$ & $77.1 \pm 5.7$ & $\textbf{76.3} \pm \textbf{5.3}$ & $\textbf{75.2} \pm \textbf{5.5}$ & $\textbf{70.9} \pm \textbf{8.6}$ & $74.5 \pm \textbf{5.7}$ \\

	\modelf & $\textbf{76.6} \pm \textbf{4.4}$ & $76.3 \pm \textbf{3.9}$ & $\textbf{78.8} \pm \textbf{3.1}$ & $73.8 \pm \textbf{5.2}$ & $\textbf{75.8} \pm 3.9$ & $\textbf{77.2} \pm \textbf{5.2}$ & $76.2 \pm \textbf{5.3}$ & $75.0 \pm 5.6$ & $69.4 \pm 9.8$ & $\textbf{74.6} \pm 6.7$ \\ 

\hline

	\oracle & $81.9 \pm 3.4$ & $80.5 \pm 3.5$ & $80.7 \pm 2.8$ & $78.0 \pm 4.5$ & $80.7 \pm 2.5$ & $83.5 \pm 3.4$ & $81.8 \pm 3.5$ & $81.4 \pm 2.9$ & $79.1 \pm 5.1$ & $79.0 \pm 3.8$ \\ 

\hline

\end{tabular}
\end{adjustbox}
\caption{Intent classification: F1 scores and standard deviations for each source and target domain pair in the UDA setup. Each number is calculated across the 5 different task labels, 25 different seeds and randomly sampled training and development sets.}
\label{tab:mantis_uda}
\end{table*}

\begin{table*}
\centering
\begin{adjustbox}{width=0.9\textwidth}

\begin{tabular}{ | l || c | c | c | c | c | c | c | c |}

\hline
& \textbf{AP $\rightarrow$ DI} & \textbf{AP $\rightarrow$ EN} & \textbf{AP $\rightarrow$ GA} & \textbf{AP $\rightarrow$ GI} & \textbf{AP $\rightarrow$ SC} & \textbf{AP $\rightarrow$ SE} & \textbf{AP $\rightarrow$ TR} & \textbf{AP $\rightarrow$ WO} \\
\hline
	\noda & $74.5 \pm 5.4$ & $69.9 \pm 6.7$ & $75.5 \pm 6.3$ & $76.7 \pm 4.6$ & $68.5 \pm 6.9$ & $76.9 \pm 4.5$ & $79.4 \pm 6.5$ & $72.6 \pm 9.2$ \\

	\dann & $74.3 \pm 5.2$ & $71.9 \pm 6.1$ & $76.6 \pm 6.2$ & $76.9 \pm 4.6$ & $71.0 \pm 7.1$ & $77.3 \pm 4.5$ & $79.4 \pm 7.0$ & $75.4 \pm 8.5$ \\

	\eda & $70.6 \pm 6.4$ & $66.1 \pm 7.0$ & $72.3 \pm 6.4$ & $73.2 \pm 5.4$ & $62.5 \pm 7.4$ & $73.2 \pm 5.3$ & $76.4 \pm 6.6$ & $68.4 \pm 9.8$ \\

	\rcrr & $74.6 \pm 5.7$ & $69.9 \pm 6.4$ & $75.9 \pm 6.2$ & $76.7 \pm 4.8$ & $68.4 \pm 7.7$ & $76.6 \pm 4.9$ & $79.5 \pm 6.5$ & $73.2 \pm 9.7$ \\

\hline
	\dcrr & $75.4 \pm 4.6$ & $70.9 \pm 6.2$ & $76.7 \pm 5.8$ & $77.9 \pm 4.3$ & $69.7 \pm 7.1$ & $78.1 \pm 4.2$ & $79.8 \pm 6.8$ & $74.3 \pm 8.5$ \\

	\rcdr & $74.8 \pm 4.9$ & $70.3 \pm 6.0$ & $75.9 \pm 5.8$ & $76.7 \pm 4.3$ & $68.8 \pm 7.1$ & $76.8 \pm 4.5$ & $78.7 \pm 6.2$ & $72.7 \pm 8.5$ \\ 

\hline

	\model & $76.1 \pm 4.0$ & $\textbf{72.1} \pm 5.6$ & $\textbf{77.8} \pm 5.4$ & $\textbf{78.6} \pm 4.0$ & $\textbf{71.2} \pm \textbf{6.1}$ & $78.4 \pm 3.9$ & $\textbf{80.6} \pm \textbf{6.0}$ & $\textbf{77.0} \pm 7.5$ \\

	\modelf & $\textbf{76.8} \pm \textbf{3.3}$ & $\textbf{72.1} \pm \textbf{4.9}$ & $77.7 \pm \textbf{4.8}$ & $78.3 \pm \textbf{3.9}$ & $70.8 \pm 6.7$ & $\textbf{78.5} \pm \textbf{3.5}$ & $80.0 \pm 6.1$ & $75.7 \pm \textbf{6.4}$ \\ 

\hline

	\oracle & $79.7 \pm 2.8$ & $77.8 \pm 3.7$ & $81.8 \pm 3.7$ & $81.4 \pm 2.6$ & $77.5 \pm 4.2$ & $82.1 \pm 2.6$ & $83.8 \pm 4.9$ & $80.3 \pm 5.5$ \\

\hline
\hline

& \textbf{DB $\rightarrow$ DI} & \textbf{DB $\rightarrow$ EN} & \textbf{DB $\rightarrow$ GA} & \textbf{DB $\rightarrow$ GI} & \textbf{DB $\rightarrow$ SC} & \textbf{DB $\rightarrow$ SE} & \textbf{DB $\rightarrow$ TR} & \textbf{DB $\rightarrow$ WO} \\
\hline
	\noda & $71.3 \pm 6.5$ & $67.0 \pm 7.2$ & $72.2 \pm 7.5$ & $73.5 \pm 4.6$ & $65.8 \pm 8.2$ & $74.2 \pm 6.1$ & $73.9 \pm 10.2$ & $69.9 \pm 10.1$ \\

	\dann & $73.6 \pm 5.5$ & $69.8 \pm 6.3$ & $74.7 \pm 6.6$ & $74.7 \pm 4.8$ & $68.5 \pm 7.7$ & $75.0 \pm 5.5$ & $76.4 \pm \textbf{7.7}$ & $71.9 \pm 8.8$ \\

	\eda & $67.8 \pm 7.8$ & $63.9 \pm 6.9$ & $69.0 \pm 7.5$ & $72.2 \pm 4.8$ & $61.1 \pm \textbf{7.2}$ & $71.4 \pm 7.0$ & $70.0 \pm 9.8$ & $65.8 \pm 9.9$ \\

	\rcrr & $72.0 \pm 7.5$ & $67.2 \pm 6.5$ & $72.7 \pm 7.2$ & $74.7 \pm 4.1$ & $65.8 \pm 8.3$ & $74.2 \pm 5.9$ & $75.0 \pm 9.6$ & $68.6 \pm 10.6$ \\ 

\hline
	\dcrr & $73.0 \pm 5.4$ & $68.6 \pm 6.4$ & $73.8 \pm 6.5$ & $74.6 \pm 4.2$ & $67.3 \pm 7.8$ & $74.9 \pm \textbf{5.0}$ & $76.3 \pm 8.7$ & $70.9 \pm 8.7$ \\

	\rcdr & $72.0 \pm 7.8$ & $66.8 \pm 7.0$ & $72.0 \pm 8.0$ & $74.5 \pm 5.2$ & $65.7 \pm 8.7$ & $74.5 \pm 6.3$ & $73.7 \pm 10.6$ & $68.6 \pm 10.8$ \\ 
\hline

	\model & $\textbf{74.3} \pm \textbf{4.9}$ & $\textbf{70.3} \pm \textbf{5.5}$ & $\textbf{75.6} \pm \textbf{6.2}$ & $\textbf{75.7} \pm \textbf{3.8}$ & $\textbf{68.8} \pm 7.9$ & $\textbf{76.1} \pm \textbf{5.0}$ & $\textbf{77.8} \pm 8.1$ & $\textbf{73.1} \pm \textbf{8.2}$ \\

	\modelf & $73.6 \pm 6.4$ & $69.8 \pm 6.4$ & $75.3 \pm 6.7$ & $75.2 \pm 4.7$ & $68.0 \pm 7.5$ & $75.9 \pm 5.4$ & $77.0 \pm 8.4$ & $71.6 \pm 10.1$ \\ 

\hline

	\oracle & $78.9 \pm 3.1$ & $75.5 \pm 4.1$ & $80.0 \pm 4.2$ & $79.7 \pm 3.4$ & $76.0 \pm 5.2$ & $80.6 \pm 3.5$ & $83.6 \pm 5.2$ & $80.0 \pm 5.4$ \\

\hline
\hline

& \textbf{EL $\rightarrow$ DI} & \textbf{EL $\rightarrow$ EN} & \textbf{EL $\rightarrow$ GA} & \textbf{EL $\rightarrow$ GI} & \textbf{EL $\rightarrow$ SC} & \textbf{EL $\rightarrow$ SE} & \textbf{EL $\rightarrow$ TR} & \textbf{EL $\rightarrow$ WO} \\
\hline
	\noda & $72.5 \pm 6.3$ & $67.2 \pm 7.9$ & $69.5 \pm 9.4$ & $71.7 \pm 7.0$ & $66.3 \pm 8.9$ & $74.0 \pm 6.7$ & $74.1 \pm 10.8$ & $70.7 \pm 10.7$ \\

	\dann & $73.2 \pm 6.3$ & $69.2 \pm 6.6$ & $72.3 \pm 8.2$ & $73.5 \pm 5.8$ & $68.1 \pm 8.3$ & $75.3 \pm 5.7$ & $76.4 \pm 9.4$ & $71.9 \pm 9.9$ \\

	\eda & $71.1 \pm 6.2$ & $64.0 \pm 6.6$ & $67.7 \pm 8.4$ & $70.3 \pm 6.1$ & $62.5 \pm 7.6$ & $72.1 \pm 5.6$ & $71.2 \pm 8.2$ & $70.4 \pm \textbf{8.7}$ \\

	\rcrr & $73.7 \pm 5.9$ & $67.5 \pm 7.8$ & $70.6 \pm 8.8$ & $73.1 \pm 6.2$ & $66.7 \pm 8.1$ & $76.0 \pm 5.8$ & $74.9 \pm 9.1$ & $70.9 \pm 9.5$ \\ 
\hline

	\dcrr & $74.4 \pm 5.1$ & $69.0 \pm 6.6$ & $71.9 \pm 8.7$ & $73.2 \pm 5.9$ & $68.2 \pm 8.2$ & $75.6 \pm 5.9$ & $76.8 \pm 9.2$ & $72.1 \pm 10.2$ \\

	\rcdr & $74.1 \pm 5.4$ & $68.1 \pm 7.3$ & $71.9 \pm 7.8$ & $73.2 \pm 5.6$ & $67.3 \pm 8.8$ & $76.7 \pm 5.7$ & $75.6 \pm 8.4$ & $71.3 \pm 8.9$ \\ 

\hline

	\model & $\textbf{75.1} \pm 5.0$ & $\textbf{70.7} \pm \textbf{5.7}$ & $\textbf{74.0} \pm \textbf{7.3}$ & $\textbf{74.7} \pm 5.3$ & $\textbf{69.1} \pm 7.4$ & $77.2 \pm 5.5$ & $\textbf{78.9} \pm 7.6$ & $\textbf{73.9} \pm 9.2$ \\

	\modelf & $74.6 \pm \textbf{4.7}$ & $69.8 \pm 5.9$ & $72.6 \pm 7.6$ & $73.9 \pm \textbf{5.0}$ & $67.8 \pm \textbf{7.3}$ & $\textbf{77.4} \pm \textbf{4.6}$ & $77.1 \pm \textbf{6.8}$ & $72.1 \pm 9.5$ \\ 

\hline

	\oracle & $77.8 \pm 3.5$ & $76.5 \pm 4.5$ & $79.2 \pm 5.0$ & $78.7 \pm 3.6$ & $76.4 \pm 5.8$ & $80.3 \pm 3.7$ & $81.5 \pm 5.3$ & $80.1 \pm 5.8$ \\

\hline
\hline

& \textbf{PH $\rightarrow$ DI} & \textbf{PH $\rightarrow$ EN} & \textbf{PH $\rightarrow$ GA} & \textbf{PH $\rightarrow$ GI} & \textbf{PH $\rightarrow$ SC} & \textbf{PH $\rightarrow$ SE} & \textbf{PH $\rightarrow$ TR} & \textbf{PH $\rightarrow$ WO} \\
\hline
	\noda & $66.5 \pm 8.9$ & $65.4 \pm 5.7$ & $64.8 \pm 8.3$ & $66.1 \pm 8.7$ & $70.2 \pm 8.8$ & $68.7 \pm 8.6$ & $64.0 \pm 9.3$ & $70.2 \pm 9.5$ \\

	\dann & $71.7 \pm 6.7$ & $69.0 \pm 5.0$ & $71.5 \pm 7.8$ & $71.4 \pm 6.0$ & $73.1 \pm 7.3$ & $75.5 \pm 5.3$ & $71.0 \pm 8.7$ & $72.9 \pm 7.7$ \\

	\eda & $64.8 \pm 7.7$ & $61.9 \pm 5.1$ & $61.8 \pm \textbf{6.0}$ & $65.8 \pm 6.1$ & $66.3 \pm 7.4$ & $66.0 \pm 6.5$ & $61.8 \pm \textbf{7.5}$ & $68.7 \pm 7.7$ \\

	\rcrr & $66.9 \pm 7.4$ & $65.7 \pm 5.8$ & $65.4 \pm 6.9$ & $67.2 \pm 6.8$ & $71.3 \pm 8.2$ & $69.9 \pm 7.1$ & $64.6 \pm 9.7$ & $69.4 \pm 8.9$ \\ 
\hline

	\dcrr & $69.9 \pm 7.5$ & $69.0 \pm 4.7$ & $68.3 \pm 8.4$ & $68.8 \pm 7.9$ & $72.1 \pm 8.1$ & $72.1 \pm 7.1$ & $68.7 \pm 9.2$ & $73.5 \pm 7.0$ \\

	\rcdr & $70.0 \pm 6.8$ & $67.7 \pm 5.3$ & $67.7 \pm 7.3$ & $69.5 \pm 6.3$ & $73.4 \pm 6.9$ & $71.7 \pm 7.4$ & $66.9 \pm 9.1$ & $73.7 \pm 7.9$ \\ 
\hline

	\model & $\textbf{73.7} \pm 5.6$ & $71.4 \pm \textbf{4.1}$ & $74.2 \pm 7.5$ & $\textbf{73.0} \pm 5.8$ & $\textbf{74.6} \pm 6.1$ & $76.7 \pm 5.0$ & $74.0 \pm \textbf{7.5}$ & $75.1 \pm \textbf{6.5}$ \\

	\modelf & $\textbf{73.7} \pm \textbf{5.1}$ & $\textbf{72.6} \pm 4.4$ & $\textbf{75.3} \pm 6.4$ & $72.6 \pm \textbf{5.5}$ & $74.5 \pm \textbf{5.5}$ & $\textbf{77.5} \pm \textbf{4.9}$ & $\textbf{74.6} \pm 7.7$ & $\textbf{75.9} \pm 7.7$ \\ 

\hline

	\oracle & $78.0 \pm 2.8$ & $76.7 \pm 3.8$ & $80.9 \pm 3.9$ & $79.4 \pm 2.8$ & $78.6 \pm 4.1$ & $81.6 \pm 2.7$ & $81.4 \pm 5.1$ & $80.9 \pm 5.0$ \\

\hline
\hline

& \textbf{ST $\rightarrow$ DI} & \textbf{ST $\rightarrow$ EN} & \textbf{ST $\rightarrow$ GA} & \textbf{ST $\rightarrow$ GI} & \textbf{ST $\rightarrow$ SC} & \textbf{ST $\rightarrow$ SE} & \textbf{ST $\rightarrow$ TR} & \textbf{ST $\rightarrow$ WO} \\
\hline
	\noda & $70.1 \pm 6.8$ & $68.5 \pm 5.2$ & $66.5 \pm 6.8$ & $73.8 \pm 5.1$ & $67.2 \pm 7.8$ & $74.5 \pm 4.5$ & $68.8 \pm 10.2$ & $70.7 \pm 7.4$ \\

	\dann & $73.5 \pm 5.3$ & $70.2 \pm 5.5$ & $70.6 \pm \textbf{4.9}$ & $74.9 \pm 4.1$ & $69.1 \pm 6.7$ & $76.7 \pm 3.8$ & $74.4 \pm 7.4$ & $73.1 \pm 6.8$ \\

	\eda & $69.6 \pm 7.0$ & $66.9 \pm 4.6$ & $66.4 \pm 8.0$ & $72.8 \pm 5.6$ & $63.8 \pm 6.2$ & $72.1 \pm 5.8$ & $70.6 \pm 10.3$ & $69.4 \pm 7.9$ \\

	\rcrr & $71.7 \pm 6.7$ & $69.5 \pm 5.1$ & $68.0 \pm 7.4$ & $74.7 \pm 4.9$ & $68.5 \pm 6.8$ & $75.2 \pm 5.1$ & $70.6 \pm 10.1$ & $71.4 \pm 7.3$ \\ 
\hline

	\dcrr & $72.4 \pm 5.6$ & $71.0 \pm 5.2$ & $67.9 \pm 6.4$ & $74.5 \pm 5.1$ & $69.9 \pm 7.2$ & $75.4 \pm 4.4$ & $73.2 \pm 8.1$ & $73.6 \pm 7.2$ \\

	\rcdr & $70.9 \pm 6.8$ & $69.5 \pm 4.5$ & $68.0 \pm 7.4$ & $74.5 \pm 5.5$ & $69.3 \pm 7.0$ & $75.4 \pm 4.8$ & $69.8 \pm 10.2$ & $73.1 \pm 7.0$ \\ 
\hline

	\model & $75.4 \pm 4.3$ & $72.5 \pm 4.7$ & $71.9 \pm 6.0$ & $76.2 \pm \textbf{3.8}$ & $70.3 \pm 5.8$ & $77.5 \pm 4.1$ & $\textbf{77.9} \pm \textbf{6.2}$ & $\textbf{75.0} \pm 6.9$ \\

	\modelf & $\textbf{76.0} \pm \textbf{3.5}$ & $\textbf{72.6} \pm \textbf{4.3}$ & $\textbf{72.8} \pm 5.1$ & $\textbf{76.6} \pm \textbf{3.8}$ & $\textbf{70.8} \pm \textbf{5.2}$ & $\textbf{77.9} \pm \textbf{2.9}$ & $77.3 \pm 6.5$ & $74.4 \pm \textbf{6.7}$ \\ 

\hline

	\oracle & $78.3 \pm 2.7$ & $75.0 \pm 3.0$ & $79.9 \pm 3.3$ & $80.1 \pm 2.5$ & $75.8 \pm 4.9$ & $80.1 \pm 2.9$ & $83.5 \pm 5.4$ & $81.0 \pm 4.3$ \\

\hline
\hline

& \textbf{UB $\rightarrow$ DI} & \textbf{UB $\rightarrow$ EN} & \textbf{UB $\rightarrow$ GA} & \textbf{UB $\rightarrow$ GI} & \textbf{UB $\rightarrow$ SC} & \textbf{UB $\rightarrow$ SE} & \textbf{UB $\rightarrow$ TR} & \textbf{UB $\rightarrow$ WO} \\
\hline
	\noda & $71.7 \pm 7.5$ & $66.6 \pm 6.9$ & $73.6 \pm 7.3$ & $72.9 \pm 5.4$ & $67.6 \pm 7.6$ & $74.2 \pm 6.0$ & $73.9 \pm 9.4$ & $68.4 \pm 9.1$ \\

	\dann & $72.1 \pm 6.6$ & $69.1 \pm 7.2$ & $74.1 \pm 6.5$ & $73.5 \pm 5.3$ & $\textbf{71.4} \pm \textbf{7.4}$ & $75.0 \pm 5.5$ & $75.5 \pm \textbf{8.1}$ & $71.3 \pm \textbf{7.9}$ \\

	\eda & $68.4 \pm 8.3$ & $63.5 \pm 6.7$ & $72.0 \pm 7.3$ & $71.7 \pm 5.6$ & $62.5 \pm 7.7$ & $71.4 \pm 6.5$ & $69.9 \pm 10.7$ & $64.6 \pm 9.8$ \\

	\rcrr & $73.1 \pm 7.1$ & $66.7 \pm 6.6$ & $74.2 \pm 7.4$ & $74.5 \pm 5.0$ & $68.7 \pm 7.9$ & $74.8 \pm 5.4$ & $73.8 \pm 10.6$ & $68.8 \pm 10.0$ \\ 

\hline

	\dcrr & $72.8 \pm 7.0$ & $67.9 \pm 6.4$ & $74.5 \pm 6.7$ & $73.9 \pm 5.1$ & $68.7 \pm 8.0$ & $74.8 \pm 5.7$ & $75.3 \pm 9.0$ & $70.4 \pm 9.5$ \\

	\rcdr & $73.4 \pm 6.7$ & $67.5 \pm 6.2$ & $74.7 \pm 6.9$ & $75.0 \pm \textbf{4.7}$ & $68.2 \pm \textbf{7.4}$ & $75.2 \pm \textbf{5.0}$ & $74.9 \pm 11.1$ & $69.5 \pm 9.8$ \\ 

\hline

	\model & $74.6 \pm 6.0$ & $\textbf{70.4} \pm \textbf{6.1}$ & $75.8 \pm \textbf{6.1}$ & $\textbf{75.6} \pm \textbf{4.7}$ & $70.7 \pm 7.5$ & $\textbf{76.3} \pm 5.4$ & $77.4 \pm \textbf{8.1}$ & $72.2 \pm 8.8$ \\

	\modelf & $\textbf{75.1} \pm \textbf{5.8}$ & $70.3 \pm \textbf{6.1}$ & $\textbf{76.3} \pm 6.2$ & $75.4 \pm 4.9$ & $70.0 \pm 7.8$ & $76.0 \pm 5.6$ & $\textbf{77.9} \pm 8.9$ & $\textbf{72.3} \pm 8.5$ \\ 

\hline

	\oracle & $79.5 \pm 3.5$ & $77.2 \pm 3.5$ & $81.9 \pm 3.5$ & $80.7 \pm 3.0$ & $78.8 \pm 4.2$ & $81.0 \pm 3.4$ & $84.2 \pm 5.0$ & $80.4 \pm 4.8$ \\ 

\hline

\end{tabular}
\end{adjustbox}
\caption{Intent classification: F1 scores and standard deviations for each source and target domain pair in the ADA setup. Each number is calculated across the 5 different task labels, 25 different seeds and randomly sampled training and development sets}
\label{tab:mantis_ada}
\end{table*}

\clearpage
\onecolumn

\end{document}